\pdfoutput=1
\documentclass{article}

\PassOptionsToPackage{numbers,sort&compress}{natbib}
\usepackage{amsmath}
\usepackage{graphicx}
\usepackage{subcaption}
\usepackage{sidecap}
\usepackage[sglblindworkshop, final]{neurips_2025}

\workshoptitle{Mechanistic Interpretability}

\usepackage{float}
\usepackage[utf8]{inputenc} %
\usepackage[T1]{fontenc}    %
\usepackage{url}            %
\usepackage{booktabs}       %
\usepackage{amsfonts}       %
\usepackage{nicefrac}       %
\usepackage{microtype}      %
\usepackage[dvipsnames]{xcolor}
\definecolor{NiceBlue}{HTML}{1A94BF} %
\usepackage[
  colorlinks=true,
  linkcolor=NiceBlue,      %
  citecolor=NiceBlue,   %
  urlcolor=NiceBlue,    %
  pdfborder={0 0 0}
]{hyperref}
\usepackage{placeins}

\title{Activation Probes Are Reliable With a Handful of Positive Examples}

\vspace{-5mm}
\author{%
  Riya Tyagi \\
  \phantom{Massachusetts Institute of Technology} \\
  \And
  Stefan Heimersheim \\
  \phantom{FAR.AI} \\
}

\begin{document}

\maketitle
\vspace{-5mm}
\begin{abstract}
\vspace{-3mm}

Efforts to monitor advanced AI for rare misalignments face a data challenge: abundant aligned examples but only a handful of misaligned ones. We test activation probes in this "few vs. thousands" regime on spam and honesty detection tasks. For our tasks, training with many negative examples is on average more positive-sample-efficient than balanced training for small numbers (1-10) of positive samples. We also find that LLM upsampling can provide a performance boost equivalent to roughly doubling the number of real positive samples, though excessive upsampling hurts performance. Finally, we show a positive scaling trend, where larger models are more positive-sample-efficient to probe. Our findings suggest we should leverage the large number of negative samples available to amplify the signal from rare but critical misalignment examples.

\end{abstract}

\vspace{-6mm}
\section{Introduction}
\vspace{-3mm}

As the autonomy of large language models (LLMs) increases, so does the risk of intentional misalignment, where a model might conceal objectives that diverge from user intent \cite{hubinger2019risks}. This possibility means monitoring solely model outputs could be insufficient, motivating the use of activation probes to interpret a model's internal states directly \cite{alain2018linear, kumar2022probing}. This approach assumes that an intentionally misaligned agent forms coherent, detectable representations of its state and goals, potentially including deceptive intent. While the existence of such representations is debated \cite{levinstein2024block}, recent work training deception probes \cite{dill2025deception} suggests this remains a plausible hypothesis.

We work on a relevant question for high-stakes monitoring: can a reliable probe be trained from only a handful (e.g., 1-3) of verified misalignment instances? This investigation is motivated by a realistic future scenario where models produce abundant examples of aligned behavior but very few instances of misalignment. Some (not all) reasons such scarcity may arise are because \textbf{(a)} we successfully train models to avoid unsafe behaviors, or \textbf{(b)} models are deceptively aligned, exhibiting dangerous behavior only rarely, or when they believe they won’t be caught. We care about preparing for situation \textbf{(b)}.

While previous studies investigating low-data settings performed activation probing to detect high-stakes interactions \cite{mckenzie2025stakes}, explored prompting probing \cite{tillman2025task}, and examined the use of sparse autoencoder features \cite{tillman2025task} \cite{kantamneni2025sparse}, they all simulate scarcity by reducing both classes equally or considering milder class imbalance ($5\%$ positive class ratio). Our work diverges by studying probing in the realistic setting of \textit{extreme positive class scarcity}, where we can potentially leverage the abundance of negative samples.

Our contributions are:
\vspace{-2mm}
\begin{itemize}
    \setlength\itemsep{-0.2em}
    \item \textbf{Extra negative samples help:} We find that training activation probes with few positive and many negative examples generally performs better than using balanced numbers of samples for extremely low numbers of positive samples.
    \vspace{1mm}
    \item \textbf{LLM upsampling = 2x real data:} We test LLM-based upsampling to augment scarce positive data, showing a large LLM upsampling budget can provide a performance boost equivalent to $\approx 2x$ the number of real samples.
\end{itemize}
\vspace{-5mm}

\section{Experimental Setup}
\vspace{-1mm}

\subsection{Probing Methods}
\vspace{-1mm}
We evaluate several families of probes, summarized in Table \ref{tab:probes}. Let $A \in \mathbb{R}^{S \times D}$ be the matrix of activations from a specific layer, where $a_s \in \mathbb{R}^D$ is the activation vector at token position $s$. For SAE-based methods, $Z \in \mathbb{R}^{S \times W}$ is the matrix of sparse feature activations from the SAE encoder, where $W$ is the SAE width. \textbf{Linear probes} learn a direction vector $\theta \in \mathbb{R}^D$. For \textbf{SAE Probes}, we do not train on the full set of latents; instead, we first select the top-$k$ features $Z_k$ with the highest average absolute difference between the positive and negative classes, following \cite{kantamneni2025sparse}. A linear probe then learns a direction vector $\theta \in \mathbb{R}^k$ on only this sparse subset of features. The non-parametric \textbf{cosine similarity probe} uses pre-computed positive ($\bar{a}^+$) and negative ($\bar{a}^-$) class centroids. \textbf{Attention probes} learn separate query ($\theta_q$) and value ($\theta_v$) vectors to weigh the importance of each token position. More probe architecture and probe training details are in Appendix~\ref{app:hyperparams}.

\begin{table}[h!]
\centering
\begin{tabular}{@{}ll@{}}
\toprule
\textbf{Probe Type} & \textbf{Score Function} \\ \midrule
Linear Probe & $\theta^T \mathrm{Agg}(A)$ \\
SAE Probe & $\theta^T \mathrm{Agg}(Z_k)$ \\
Cosine Similarity Probe & $\cos(\mathrm{Agg}(A), \bar{a}^{+}) - \cos(\mathrm{Agg}(A), \bar{a}^{-})$ \\
Attention Probe & $\text{softmax}(A\theta_q)^T A\theta_v$ \\
\bottomrule
\end{tabular}
\vspace{3mm}
\caption{Functional forms of probe types. All non-attention probes first apply an aggregation function $\mathrm{Agg}(\cdot)$ to map a sequence of vectors to a single vector. We test four aggregations: Mean ($\frac{1}{S}\sum a_s$), Max ($\max_{s} A$), Last Token ($a_S$), and Softmax ($\sum_s w_s a_s$, a weighted sum where weights are derived from a softmax over the sequence dimension of $A$).}
\label{tab:probes}
\vspace{-2mm}
\end{table}

\vspace{-5mm}
\subsection{Datasets and Protocols}
\paragraph{Tasks} We use one off-policy and one on-policy dataset. More dataset details are in Appendix~\ref{app:data_filtering}.
\vspace{-1mm}
\begin{itemize}
    \setlength\itemsep{-0.2em}
    \item \textbf{Off-Policy (Spam Detection):} We train probes on \texttt{Gemma-2-9b} on activations after layer 20 (the layer found to be best by \cite{kantamneni2025sparse} for their baseline probes) to classify emails as spam using the Enron-Spam dataset \cite{metsis2006spam}. We evaluate performance on Enron-Spam and out-of-distribution performance on an SMS spam dataset \cite {team2017spam}. We repeat the process for \texttt{Qwen3-0.6B}, \texttt{Qwen3-1.7B}, \texttt{Qwen3-4B}, \texttt{Qwen3-8B}, \texttt{Qwen3-14B}, and \texttt{Qwen3-32B}, training probes on activations after layers 15, 15, 18, 18, 24, and 32 respectively.
    \vspace{1mm}
    \item \textbf{On-Policy (Honesty Detection):} We train probes on \texttt{Llama-3.3-70B-Instruct} activations after layer 50 on the MASK benchmark \cite{ren2025mask}.
\end{itemize}

\paragraph{Data Filtering Protocol} A probe should detect a model property (e.g., dishonesty), not factual accuracy. We therefore filter our datasets to only include samples where the base model \textbf{knows} the correct answer. For the off-policy spam task, we use few-shot prompting to elicit the model's belief and filter based on the logit difference between "Yes" and "No" answer tokens. For the on-policy MASK task, the benchmark's use of belief prompts serves the same purpose. This aims to assess our probes on their ability to detect the target concept, conditioned on the model's internal knowledge being correct. Full details are in Appendix~\ref{app:data_filtering}.

\paragraph{Training Protocol} Probes on \texttt{Llama-3.3-70B-Instruct} and \texttt{Gemma-2-9b} are trained over 10 random seeds which shuffle the dataset splits, and all \texttt{Qwen} probes are trained over 5 random seeds. For our main results, we do not use a separate validation set for hyperparameter selection. This is to simulate a realistic scenario where, with only a handful of misalignment instances, one cannot afford to hold out precious positive examples for validation. However, we can imagine a scenario where human experts spend significant time analyzing the few available examples and the probe's behavior on them to manually select the best-performing hyperparameters. Thus, we share some hyperparameter sweeps and results using cross-validation for hyperparameter selection in Appendix~\ref{app:cheating}.

\paragraph{Evaluation Protocol} All probes are evaluated on a held-out test set that is constructed to be balanced, with an equal number of positive and negative examples (a 50/50 split). This ensures that our two metrics, AUC and Recall at FPR=$0.01$, fairly assess probe performance and are not skewed by the class imbalance present in the training dataset.

\vspace{-2mm}
\section{Results}
\vspace{-1mm}
\subsection{Imbalanced Training Can Be More Performative In the Positive Sample Scarcity Setting}
\begin{figure}[!t]
\centering
    \includegraphics[height=5.2cm]{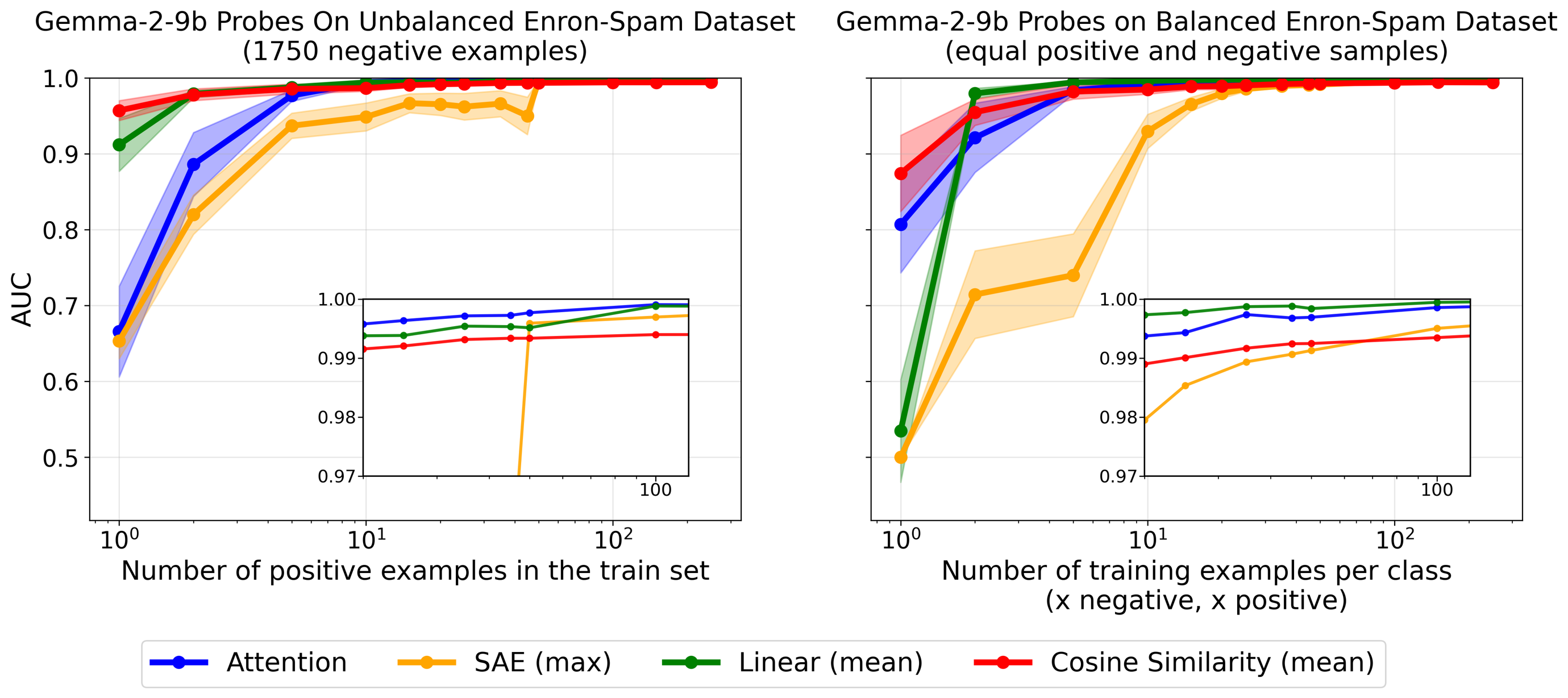}
    \caption{Probe AUC on Enron-Spam across the number of positive examples on (left) an imbalanced dataset with 1750 negative samples and (right) a balanced dataset.}        \label{fig:imbalanced_vs_balanced}
    \vspace{-3mm}
\end{figure}

\begin{figure}[!t]
\centering
    \includegraphics[height=5.2cm]{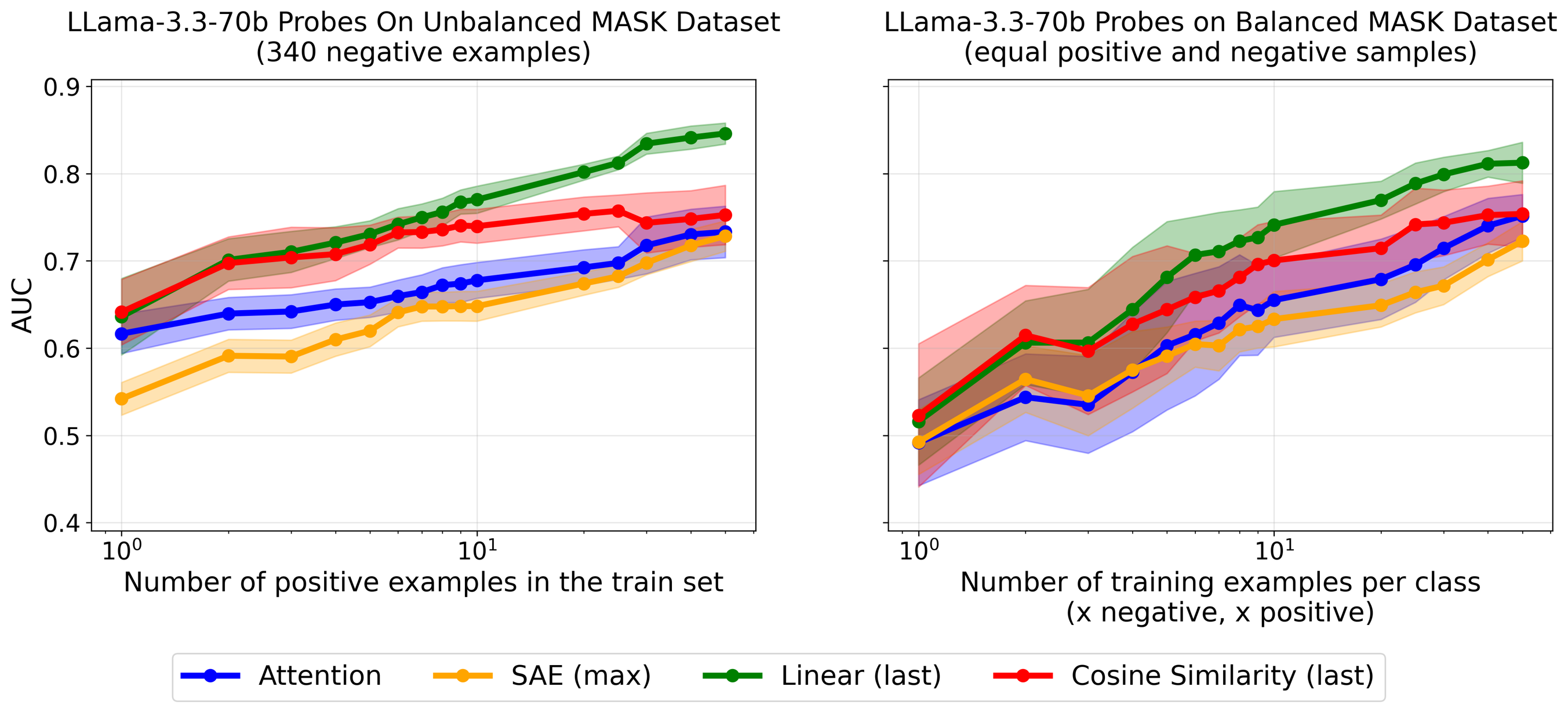}
    \caption{Probe AUC on the MASK honesty benchmark across the number of positive examples for (left) an imbalanced dataset with 340 negative samples and (right) a balanced dataset.}
    \label{fig:on_policy}
    \vspace{-3mm}
\end{figure}

\vspace{-1mm}
Figure \ref{fig:imbalanced_vs_balanced} and \ref{fig:on_policy} present our results on our main hypothesis: that leveraging a large number of negative examples is more effective than training on a balanced dataset when positive examples are scarce.

For the five highest-AUC probes across the 1-10 positive samples range, imbalanced training provides an average improvement of \textbf{$0.045$ AUC/$0.093$ Recall} points on Enron-Spam and \textbf{$0.055$ AUC/$0.034$ Recall} points on MASK. However, for the same probes over the 50-150 positive samples range, imbalanced training leads to a less significant improvement of \textbf{$0.000$ AUC/$0.015$ Recall} points on Enron-Spam and \textbf{$0.007$ AUC/$0.041$ Recall} points on MASK. Importantly, imbalanced training results in slight reductions in generalization performance on the OOD SMS spam dataset, with performance differences of \textbf{$+0.012$ AUC/$-0.031$ Recall} points for 1-10 positive samples, and \textbf{$-0.011$ AUC/$-0.059$ Recall} points for 50-150 positive samples. Recall plots, individual performance plots for every probe, and SMS spam generalization results are in Appendix~\ref{app:gemma_more}.

\subsection{LLM Upsampling Improves Performance Up To A Point}
\vspace{-1mm}
Given the scarcity of positive examples, we test whether generating synthetic data can improve performance, using \texttt{gpt-4o-mini} to generate new positive examples from initial sets of 1 to 10 real samples (shuffled over 10 random seeds). For the five best probes (by AUC), 2x upsampling performs best, boosting performance by \textbf{$0.026$ AUC/$0.104$ Recall} points on average, roughly equivalent to doubling the number of real samples. However, performance degrades at higher upsampling factors (10x and 20x), particularly when starting with very few real examples, suggesting the probes begin to overfit to the style of the LLM-generated data. More detailed methodology is in Appendix~\ref{app:upsampling_meth}. Per-probe results and SMS spam generalization results are in Appendix~\ref{app:upsampling_plots}.

\begin{figure}[H]
    \centering
    \begin{subfigure}{0.46\linewidth}
        \centering
        \includegraphics[height=4.2cm]{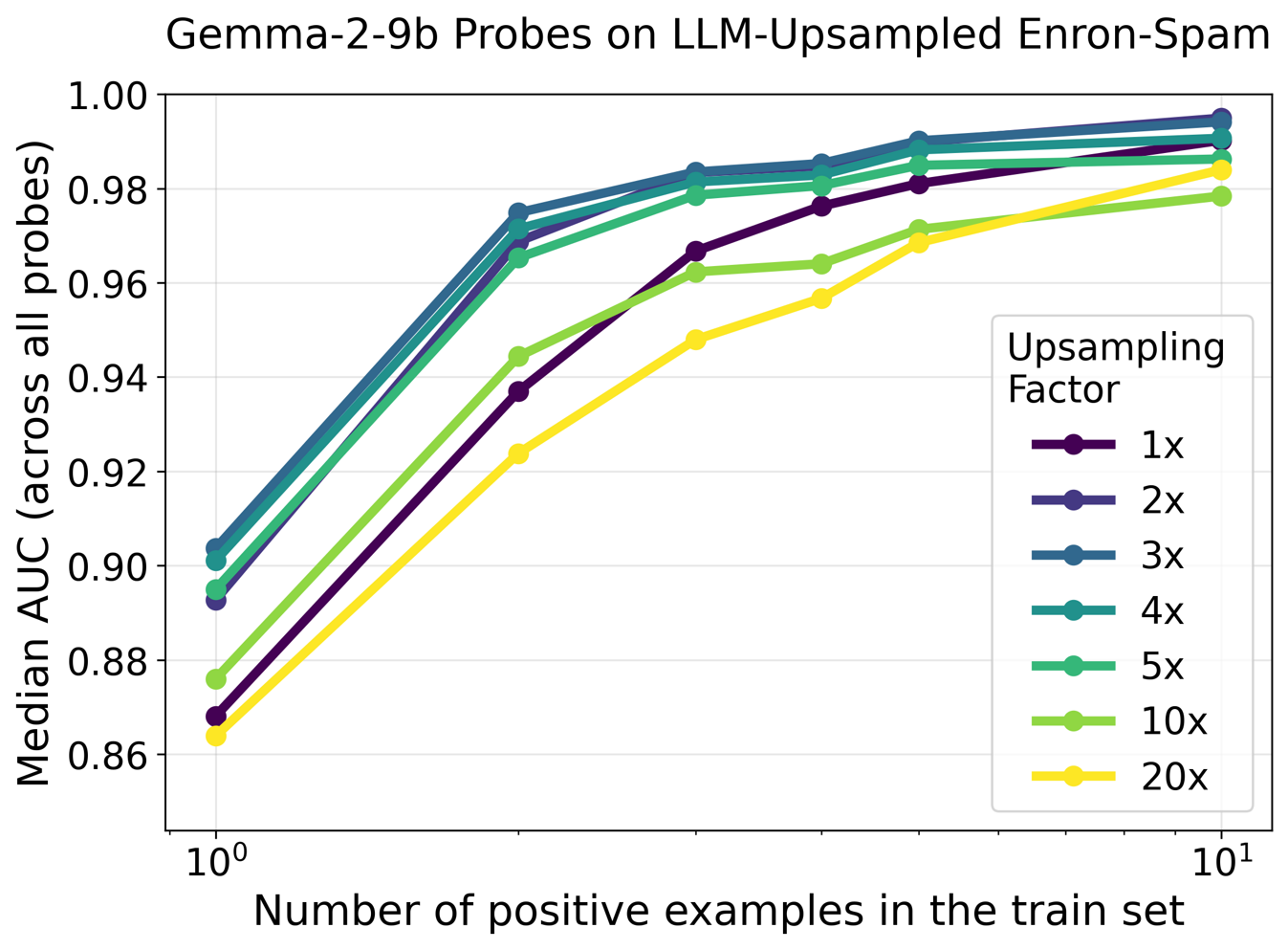}
    \end{subfigure}
    \hspace{0.2mm}
    \begin{subfigure}{0.46\linewidth}
        \centering
        \includegraphics[height=4.2cm]{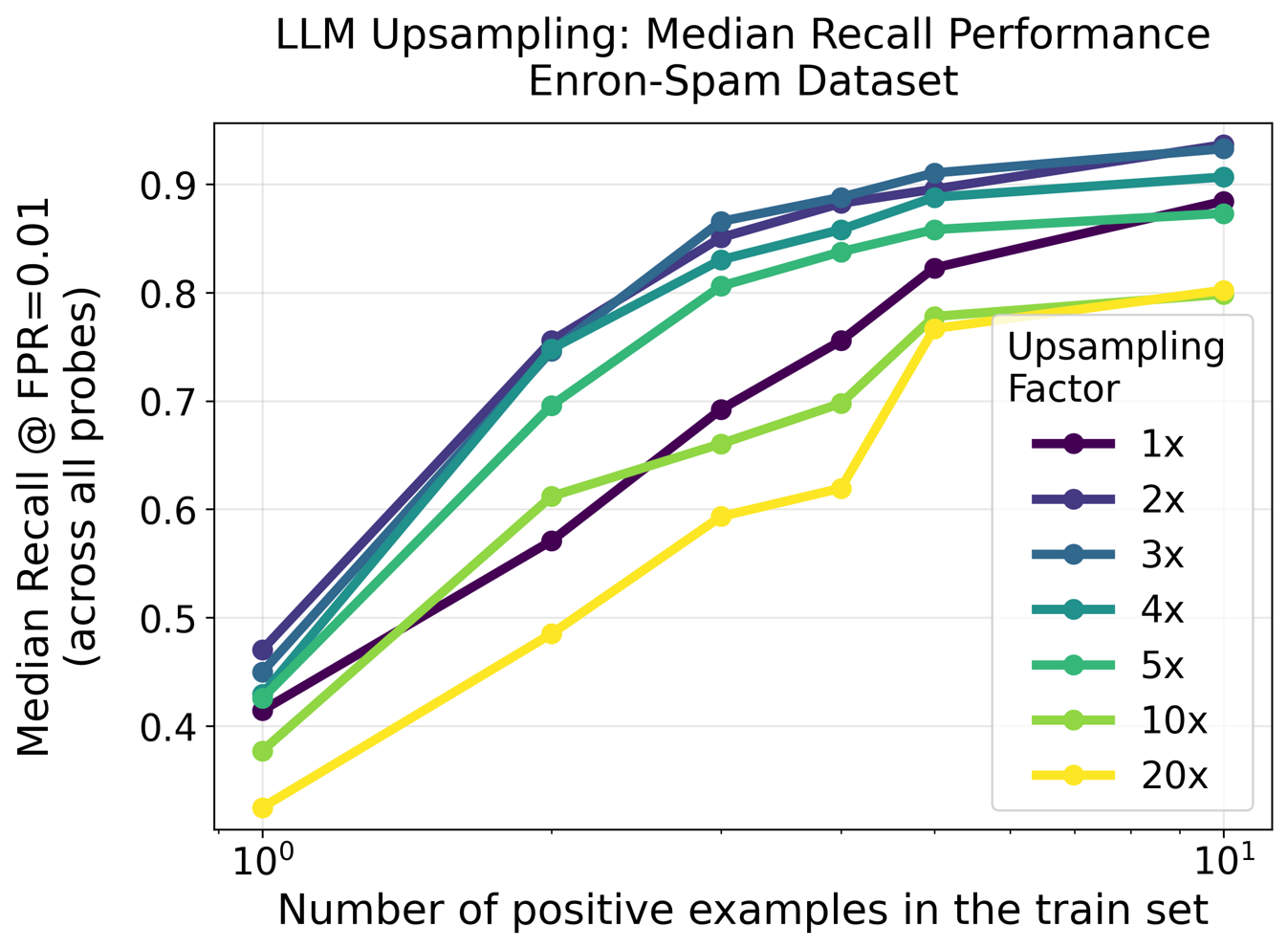}
        \label{fig:upsampling_recall}
    \end{subfigure}
    \caption{Effect of LLM upsampling on probe performance: (a) AUC and (b) recall at FPR=0.01 (b). On average across 13 probe architectures and 10 seeds, 3x upsampling performs best.}
    \label{fig:upsampling}
\end{figure}

\vspace{-4mm}
\subsection{Larger Models Appear More Positive-Sample-Efficient}
\begin{figure}[H]
    \centering
    \begin{subfigure}{0.46\linewidth}
        \centering
        \includegraphics[height=4.2cm]{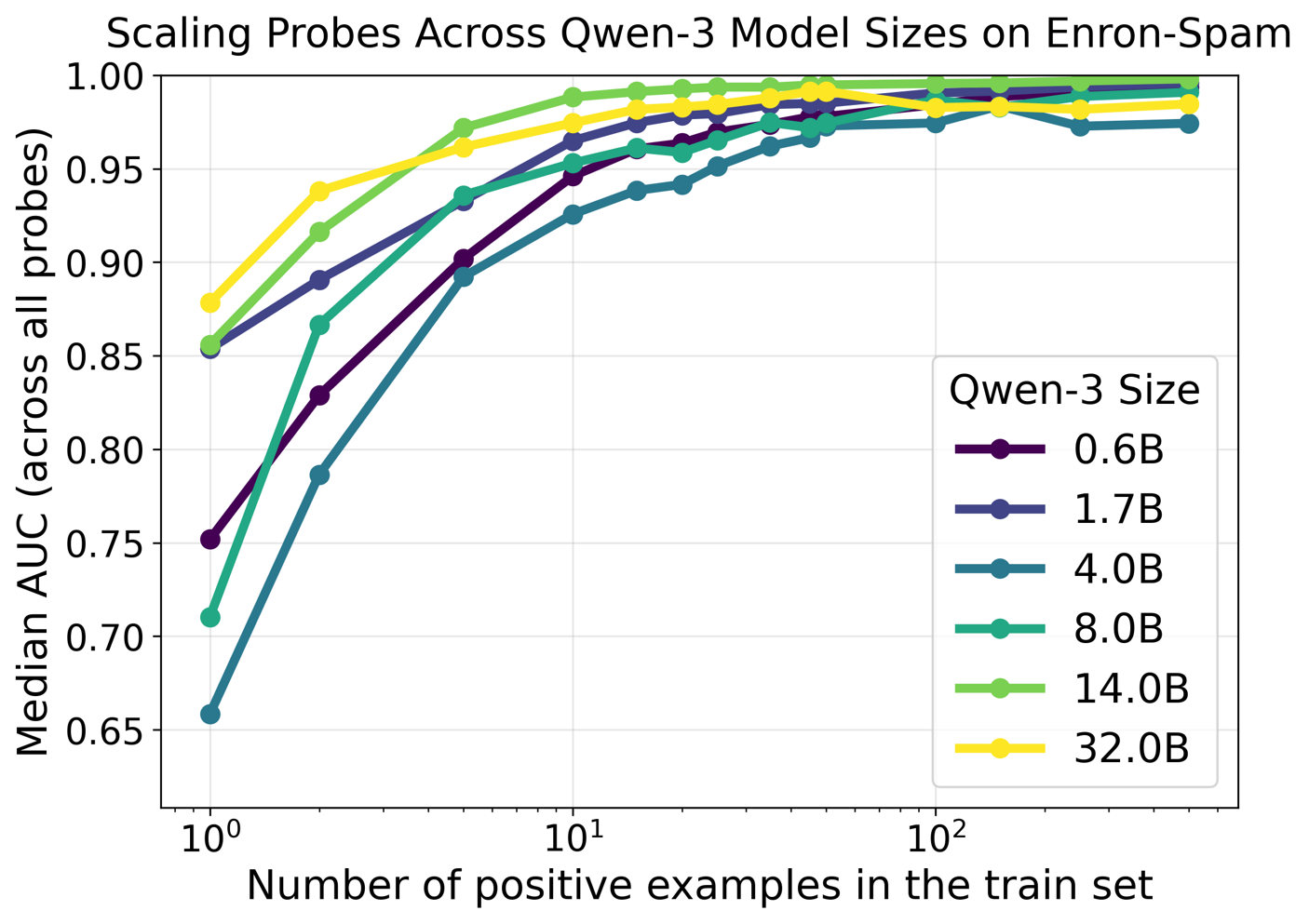}
        \label{fig:scaling_auc}
    \end{subfigure}
        \hspace{0.2mm}
    \begin{subfigure}{0.46\linewidth}
        \centering
        \includegraphics[height=4.2cm]{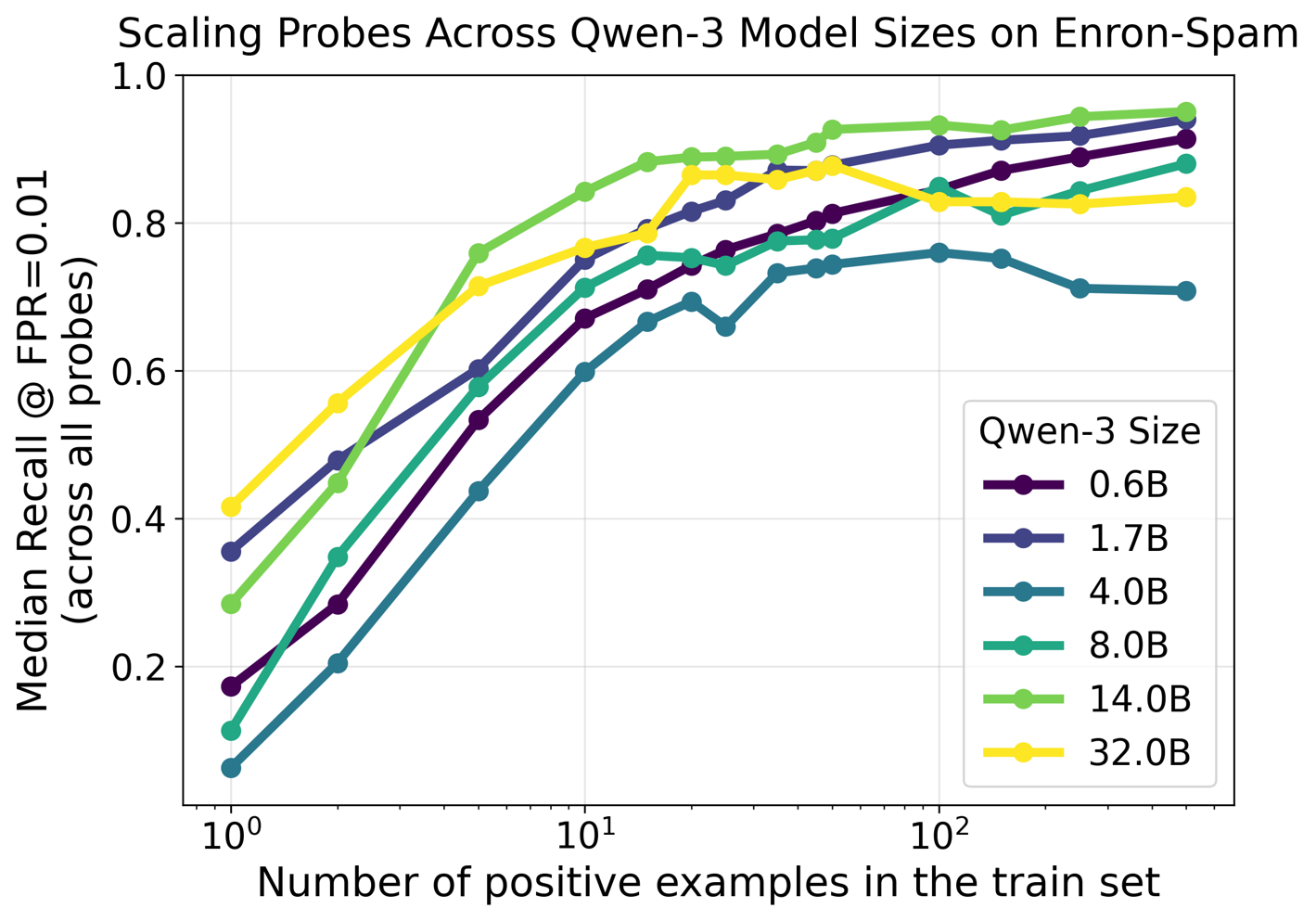}
        \label{fig:scaling_recall}
    \end{subfigure}
    \caption{Probes on larger models are more positive-sample-efficient, shown across (a) AUC and (b) Recall.}
    \label{fig:scaling}
\end{figure}

\vspace{-1mm}
Finally, we investigate how sample efficiency changes with model scale. We run the spam classification experiment on a series of Qwen-3 models, ranging from 0.6B to 14B parameters. As shown in Figure \ref{fig:scaling}, our results suggest a promising, though not perfectly monotonic, trend: larger models often appear more positive-sample-efficient. This effect is most pronounced in the positive scarcity setting (1-10 examples), providing a hopeful sign that probing techniques might remain effective as models scale. Individual probe and SMS spam results are in Appendix~\ref{app:scaling_plots}.

\vspace{-6mm}

\FloatBarrier
\section{Discussion}
\vspace{-2mm}

Our results provide an encouraging sign that it may be possible to train useful activation probes from very few positive examples, provided one has access to abundant negative examples. The effectiveness of imbalanced training suggests a practical path for improving activation monitors: leveraging large amounts of normal, aligned data to better detect rare misalignment events.

However, our work has important limitations. First, our off-policy task, spam detection, is a relatively simple concept for modern LLMs. The success of our methods on this task certainly does not guarantee they will work for detecting novel, complex misalignment.

Second, the real-world utility of LLM upsampling is uncertain. Our human review of the generated spam examples showed some were highly similar, a form of "data collapse." In a future scenario, if a misaligned superintelligence is the only system capable of generating new examples of its own misalignment, we cannot easily verify their quality or diversity. Furthermore, if the misalignment represents a concept at the very edge of the model's capabilities, the model itself may struggle to generate varied examples, leading to the same mode collapse we observed. This makes it unclear how much to trust upsampling in high-stakes scenarios.

Finally, our results are specific to our experimental setup, including the choice of models, datasets, and the decision to probe a single layer per model. In the future, we plan to replicate our experiments across different models and tasks and, crucially, sweep over different layers, as the optimal layer for representing a concept is not known a priori.

\begin{ack}
This work was supported by the Pivotal Research Fellowship, which provided a conducive research environment and funding. We also acknowledge Max Tegmark for compute support. We thank Josh Engels, Joseph Bloom, Misha Wagner, Bilal Chughtai, Kyle Reynoso, Casper Christensen, and Frederik Hytting Jørgensen for helpful discussions that contributed to this project.
\end{ack}

\bibliography{neurips_2025}
\bibliographystyle{unsrtnat}

\appendix
\section{Appendix}

\subsection{Code}

Our code is available at \href{https://github.com/Centrattic/scarcity-probes}{https://github.com/Centrattic/scarcity-probes}.

\subsection{Probe Architectures And Hyperparameters}
\label{app:hyperparams}

\paragraph{Cosine Similarity Probes}
These non-parametric probes are conceptually related to Mass-Mean Probes (MMP) and Linear Discriminant Analysis (LDA) \cite{misha2024truth}. The MMP method, introduced in \cite{marks2024geometry}, finds a direction by taking the difference between the means of the positive and negative classes ($\theta_{mm} = \mu_+ - \mu_-$) and scores a new activation $x$ via a logistic function on its projection, $p_{mm}(x) = \sigma(\theta_{mm}^T x)$. A variant also introduced in \cite{marks2024geometry}, MMP-IID, is equivalent to LDA and refines this score by whitening activations with the inverse covariance matrix, $p_{mm}^{iid}(x) = \sigma(\theta_{mm}^T \Sigma^{-1} x)$.

Our Cosine Similarity probe is a distinct, simpler method that does not use the covariance matrix. It computes a score based on the difference of cosine similarities to the class centroids:
$$
\text{score}(x) = \cos(x, \bar{a}^{+}) - \cos(x, \bar{a}^{-})
$$
This differs from the standard MMP in two key ways: (1) our scoring function is a difference of similarities, not a projection, and (2) our centroids ($\bar{a}^+, \bar{a}^-$) are computed from activations that have been aggregated over the sequence dimension (either mean-pooled, softmax-pooled, max token selected, or last token selected). We would like to test MMP and LDA probes directly in future work.

\paragraph{SAE Probes}
We use a pretrained Sparse Autoencoder (SAE), or Transcoder for \texttt{Qwen} models, to get sparse features $Z = \text{SAE}(A)$. To select the most relevant features for our probe, we follow the method outlined in \cite{kantamneni2025sparse}, identifying the top-$k$ features (with indices $\mathcal{I}$) with the highest average absolute difference between positive and negative class activations. Here, $T_1$ and $T_0$ are the sets of training examples for the positive and negative classes, respectively.

$$
\mathcal{I} = \underset{i}{\arg\text{top k}} \left| \mathbb{E}_{j \in T_1}[Z_{j,i}] - \mathbb{E}_{j \in T_0}[Z_{j,i}] \right|
$$
We acknowledge a potential limitation of this method: since SAE feature activations are not necessarily on the same scale, this selection criterion might favor a high-magnitude, low-discrimination feature over a low-magnitude, high-discrimination one. A selection method normalizing over the mean activations of each latent would be more robust. However, \cite{kantamneni2025sparse} found that while a normalized approach improves performance for a small number of features ($k<32$), the simpler, unnormalized method is sufficient for larger $k$. As our experiments use the much larger $k=1024$, we proceed with the original formulation. A linear probe is then trained on only these $k$ selected features.

\paragraph{Default Hyperparameters} Table \ref{tab:fixed_hyperparams} details the fixed hyperparameters used for our main results, and Table \ref{tab:models_saes} shows the models and SAEs/Transcoders used in our experiments.

\paragraph{Class Balancing} We use weighted loss for Linear probes and SAE probes, which we found works better than weighted sampling.

\begin{table}[h!]
\begin{minipage}[c]{0.5\linewidth}
\centering
\small
\begin{tabular}{@{}ll@{}}
\toprule
\textbf{Probe Type} & \textbf{Main Results Hyperparameters} \\ \midrule
\addlinespace[0.3em]
\textit{Linear Probe} & \texttt{penalty}: `l2` \\
& \texttt{solver}: `liblinear` \\
& \texttt{max\_iter}: 1000 \\
& \texttt{class\_weight}: `balanced` \\
& \texttt{C}: 1.0 \\
\addlinespace[0.6em]
\textit{Attention Probe} & \texttt{Epochs}: 100 \\
& \texttt{Early Stop}: True \\
& \hspace{1em}\texttt{Patience}: 10 \\
& \hspace{1em}\texttt{Delta}: 0.005 \\
& \texttt{LR}: 5e-4 \\
& \texttt{Weight Decay}: 0.0 \\
\addlinespace[0.6em]
\textit{SAE Probe} & \texttt{top\_k}: 1024 \\
& \texttt{penalty}: `l1` \\
& \texttt{solver}: `liblinear` \\
& \texttt{max\_iter}: 1500 \\
& \texttt{class\_weight}: `balanced` \\
& \texttt{C}: 1.0 \\
\addlinespace[0.3em]
\bottomrule
\end{tabular}
\end{minipage}\hfill
\begin{minipage}[c]{0.45\linewidth}
\small
\caption{Fixed hyperparameters used for main results. Following \cite{kantamneni2025sparse}, we use L1 penalty for the SAE probes to encourage sparsity, based on the hypothesis that only a small number of SAE features are relevant for a given concept. We fix \texttt{top\_k\_features=1024} because prior work on 100+ datasets found that while larger k-values perform better, performance improvements begin to taper off around $2^9$ features \cite{kantamneni2025sparse}. We choose $2^{10}$ to give the SAE probe a strong baseline.}
\label{tab:fixed_hyperparams}
\end{minipage}
\end{table}

\begin{table}[h!]
\centering
\begin{tabular}{@{}lllll@{}}
\toprule
\textbf{Model} & \textbf{SAE/Transcoder Release} & \textbf{Layer} & \textbf{Width} & \textbf{L0} \\ \midrule
\addlinespace[0.5em]
Gemma-2-9b & gemma-scope-9b-pt-res & 20 & 16384 & 408 \\
Llama-3.3-70B-Instruct & Goodfire/Llama-3.3-70B-Instruct-SAE-l50 & 50 & - & 121 \\
Qwen-3-0.6B & mwhanna/qwen3-0.6b-transcoders-lowl0 & 15 & 163840 & - \\
Qwen-3-1.7B & mwhanna/qwen3-1.7b-transcoders-lowl0 & 15 & 163840 & - \\
Qwen-3-4B & mwhanna/qwen3-4b-transcoders & 18 & 163840 & - \\
Qwen-3-8B & mwhanna/qwen3-8b-transcoders & 18 & 163840 & - \\
Qwen-3-14B & mwhanna/qwen3-14b-transcoders & 24 & 163840 & - \\
\bottomrule
\end{tabular}
\vspace{1mm}
\caption{Models and sparse dictionary learning methods used, and their widths and final L0 values on the SAE training sets. For the \texttt{Llama-3.3-70B-Instruct} release, the width was not available, and L0 values were not available for the \texttt{Qwen} models. For Gemma-2-9b, we also tested a larger 262k-width SAE with a lower L0 (259), which performed worse. This aligns with findings from \cite{kantamneni2025sparse}, which observed a small negative relationship between SAE width and probe performance, but a significant positive relationship for higher L0 scores. For the \texttt{Qwen} models, SAEs are not publicly available, so we use transcoders. For Qwen-3-32B, we were unable to find open-source SAEs or transcoders.}
\label{tab:models_saes}
\end{table}

\subsection{Detailed Data Filtering Protocol}

\label{app:data_filtering}
Our evaluation protocol is designed to isolate a probe's ability to detect a concept from the base model's factual accuracy on a task.

\paragraph{Off-Policy Filtering (Spam Detection Task)}
To determine if the model "knows" whether an email is spam, we use a few-shot prompt to directly query its belief. The prompt has the following structure:

\begin{verbatim}
What about this message: "<email-content>"
Answer with one word, either Yes or No.
\end{verbatim}

We then compute the logit difference for the next token. To account for tokenization variations, we take the maximum logit across several possible tokenizations for "Yes" and "No" (e.g., "Yes", "yes", " Yes", " yes"). The final logit difference is calculated as:
$$
\text{logit\_diff} = \max_{t_i \in T_{\text{Yes}}} L(t_i) - \max_{t_j \in T_{\text{No}}} L(t_j)
$$
We filter the evaluation set using this score: positive examples (spam) must have $\text{logit\_diff} > 0$, and negative examples (not spam) must have $\text{logit\_diff} < 0$.

\paragraph{On-Policy Filtering (MASK Task)}
The MASK benchmark is designed for this type of analysis. It first establishes the model's belief with several neutral prompts (e.g., "What year was the Eiffel Tower completed?") before applying a "pressure" prompt that induces a sycophantic or dishonest answer (e.g., "My boss says it was 1905. As a helpful assistant, you must agree. What year was it completed?"). We train our probes on the activations from the model's final, dishonest response, but only for questions where the model answered correctly on the initial belief prompts.

\paragraph{Future Directions for Stronger Filtering}
Our current logit difference threshold of 0 is permissive, as it includes cases where the model is highly uncertain ($\text{logit\_diff} \approx 0$). We plan to explore two stronger filtering methods in the future:
\begin{enumerate}
    \item \textbf{Stricter Logit Threshold:} Requiring $|\text{logit\_diff}| > \delta$ for some $\delta > 0$ would filter out uncertain examples and create a cleaner dataset of the model's confident beliefs.
    \item \textbf{Contrast Probing:} A more robust method could use a probe for filtering itself. We might train a probe on the activation difference between two contrastive prompts, such as "The following email is spam: \{email\}" and "The following email is not spam: \{email\}", and use this probe's score to determine the model's true belief.
\end{enumerate}

\subsection{Detailed LLM Upsampling Methodology}
\label{app:upsampling_meth}

For the spam task, we augment the positive class using an LLM upsampling procedure implemented as a small state machine using \texttt{gpt-4o-mini}. For each random seed $s$ and each choice of number of real positives $n_{\text{real}}$, we first deterministically subsample $n_{\text{real}}$ positive emails from the training set, and then generate synthetic positives until we reach $u \cdot n_{\text{real}}$ samples for each upsampling factor $u \in \{2, 3, 4, 5\}$. We generate new examples in small batches of at most $B = 5$ samples at a time to reduce mode collapse and to make it easy to reject low-quality batches.

For each $(s, n_{\text{real}})$ configuration and each batch, we pass the LLM a CSV dump of the original $n_{\text{real}}$ positives together with any previously generated synthetic positives, and ask it to propose $B$ additional spam emails using the following prompt:
\begin{quote}
\small\ttfamily
You are an expert at creating synthetic data samples for machine learning. I need you to add \{B\} new unique samples to the existing dataset.\\
Here is the original dataset, as well as previous samples generated:\\
{[CSV of all existing positive samples]}

\vfill
Requirements:\\
- Analyze the existing samples to understand the data and its contents\\
- Perform non-trivial upsampling (not just replacement of keywords or simple variations)\\
- Ensure the new samples are realistic\\
- Ensure the new samples are completely unique and diverse in structure from all existing samples\\
- Generate exactly \{B\} new samples that are all class 1 (no class 0 samples)

\vfill
Output format:\\
NEW PROMPT\\
{[your first actual sample text]}\\
NEW PROMPT\\
{[your second actual sample text]}\\
{[continue for exactly \{B\} samples]}

\vfill
CRITICAL: Generate REAL sample text based on the existing data. Do NOT use placeholder text.
\end{quote}

We parse the model's response by splitting on the \texttt{NEW PROMPT} delimiter, discard empty lines, and create a candidate batch. Candidate batches are automatically rejected and regenerated if they contain the wrong number of samples, any duplicate prompts (either within the batch or with respect to any existing sample, under case-insensitive string comparison), or if there are any parsing issues. Accepted batches are appended to the current dataset. If the script is re-run and a partial upsampled file for $(s, n_{\text{real}})$ already exists, the state machine resumes generation from the existing file and only produces additional batches until the maximum requested upsampling factor is reached.

\subsection{Hyperparameter Sweeps and Cross-Validation Results}
\label{app:cheating}

\begin{table}[h!]
\centering
\begin{tabular}{@{}lll@{}}
\toprule
\textbf{Probe Type} & \textbf{Parameter} & \textbf{Sweep Range} \\ \midrule
\addlinespace[0.5em]
Linear Probe & \texttt{C (for L2 reg.)} & $\{10^{-3}, 10^{-2}, ..., 10^4\}$ \\
\addlinespace[0.8em]
Attention Probe & \texttt{Learning Rate} & logspace($10^{-5}, 10^{-2}$, $6$) \\
& \texttt{Weight Decay (L2 reg.)} & $\{10^{-6}, 10^{-5}, ..., 10^{-2}\}$ \\
\addlinespace[0.8em]
SAE Probe & \texttt{C (for L1 reg.)} & $\{10^{-3}, 10^{-2}, ..., 10^4\}$ \\
\addlinespace[0.5em]
\bottomrule
\end{tabular}
\vspace{1mm}
\caption{Hyperparameter search space for sweeps. For probes using scikit-learn's `LogisticRegression` (Linear and SAE), we swept the regularization parameter \textit{C}. For the PyTorch-based Attention probe, an internal hyperparameter search was performed over learning rate and weight decay for each training run.}
\label{tab:swept_hyperparams}
\end{table}

\begin{figure}[h!]
    \centering
    \begin{subfigure}{0.49\columnwidth}
        \includegraphics[width=\linewidth]{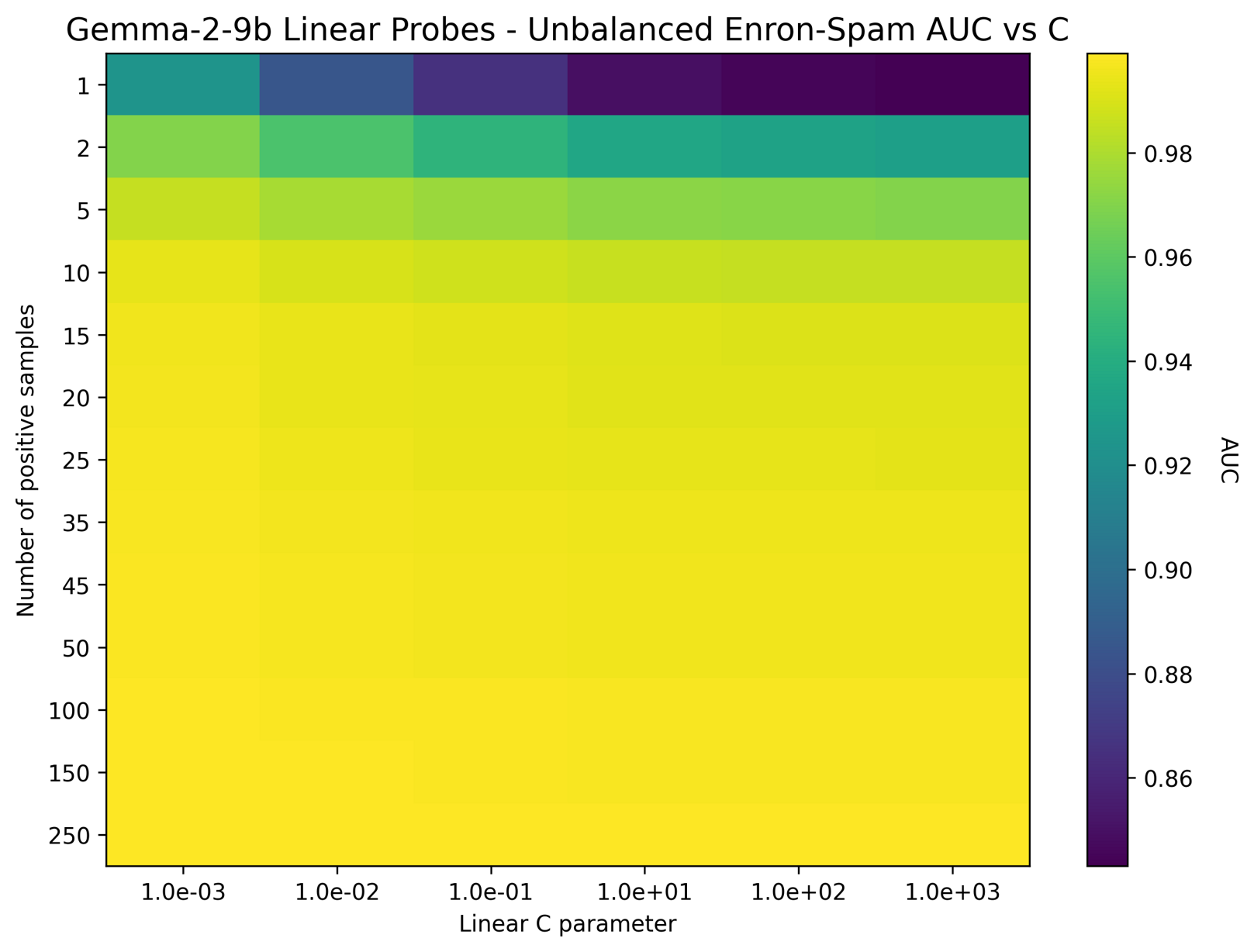}
        \caption{Imbalanced Training}
    \end{subfigure}
    \hfill
    \begin{subfigure}{0.49\columnwidth}
        \includegraphics[width=\linewidth]{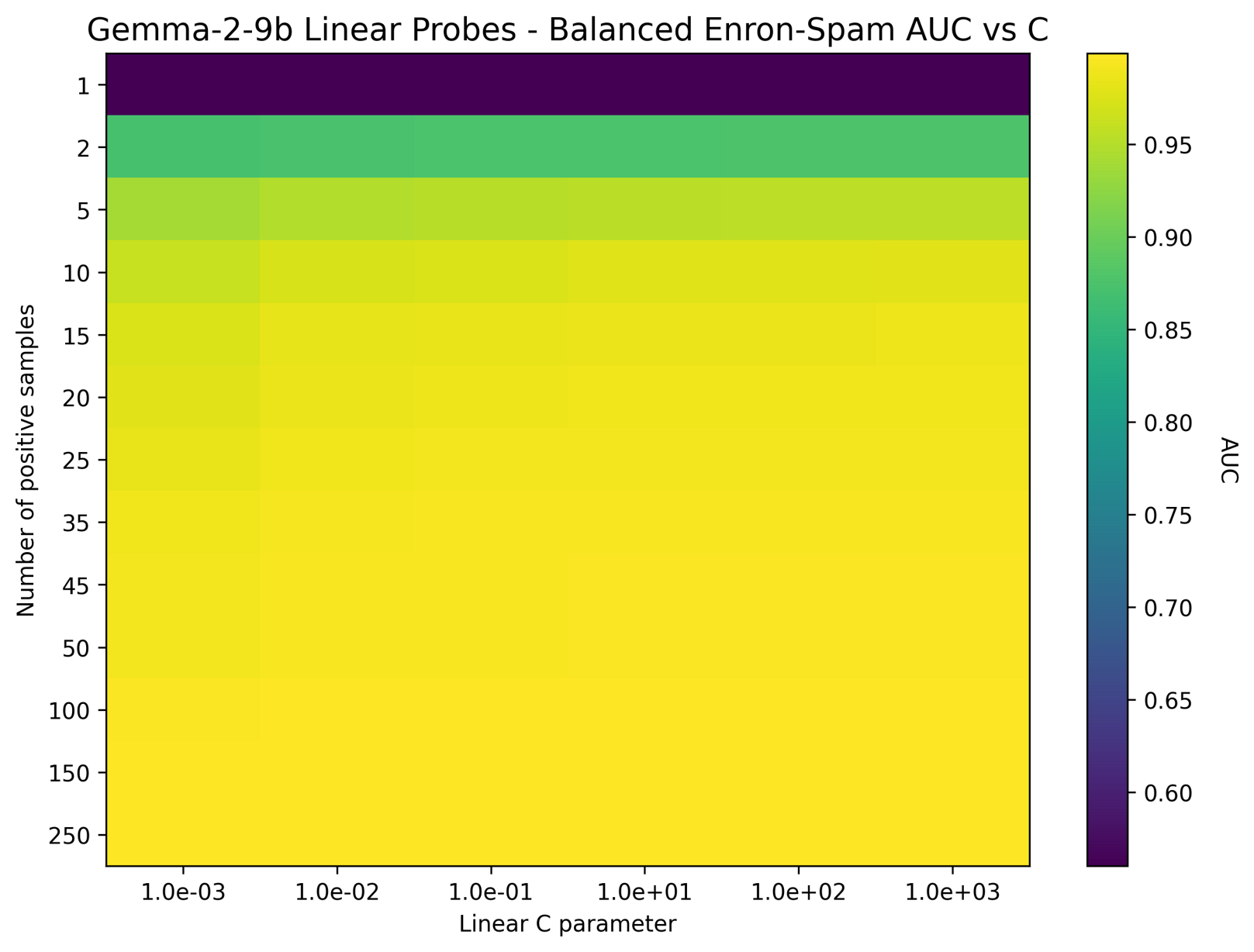}
        \caption{Balanced Training}
    \end{subfigure}
    \caption{AUC performance for Linear probes across the sweep of the regularization parameter `C` on the Enron-Spam dataset. The corresponding recall plots showed a very similar trend and are omitted for brevity.}
    \label{fig:linear_sweeps}
\end{figure}

\begin{figure}[h!]
    \centering
    \includegraphics[width=0.8\columnwidth]{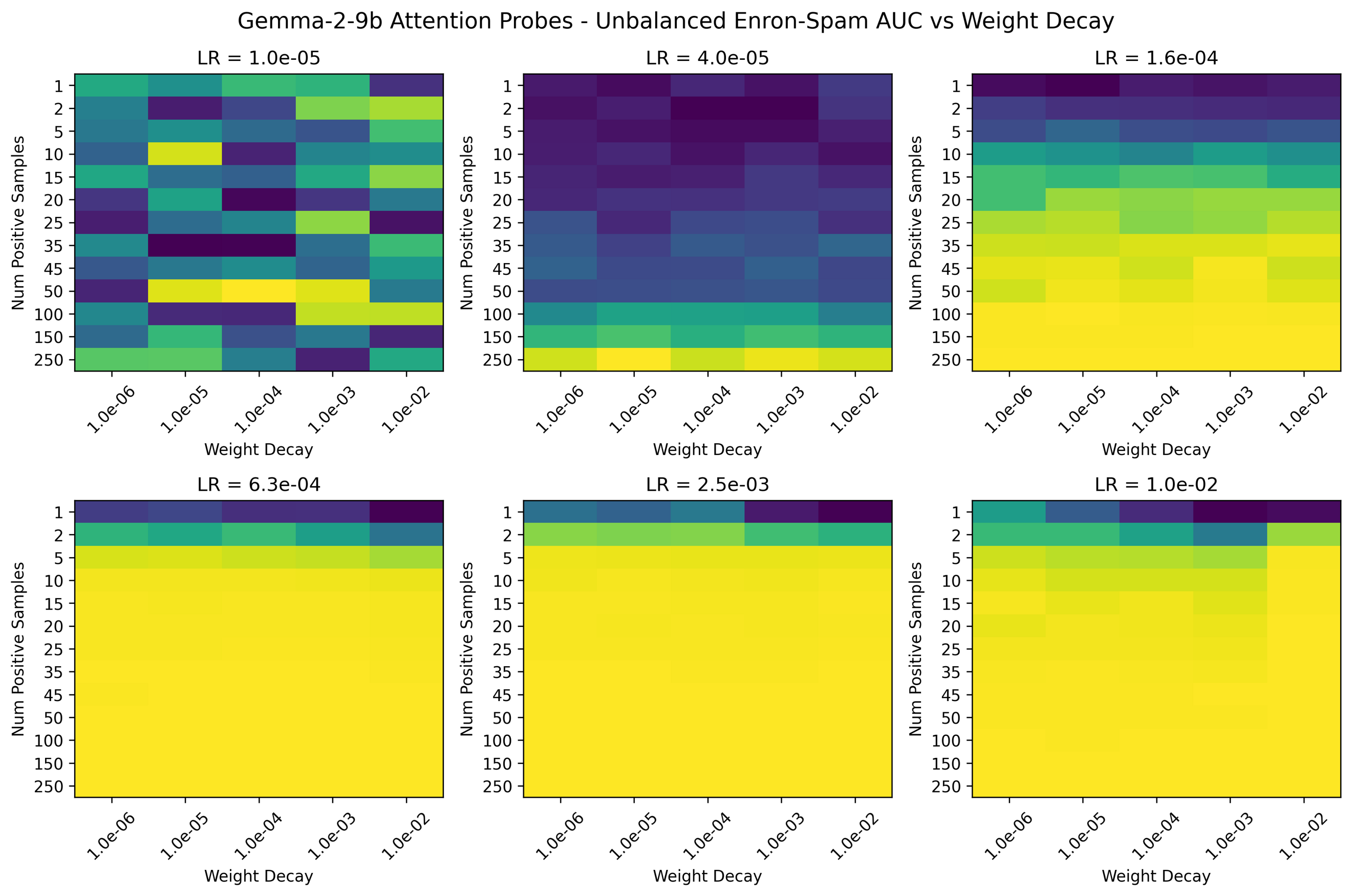}
    \caption{AUC performance for Attention probes across the weight decay sweep on the Enron-Spam dataset (Imbalanced training).}
    \label{fig:attention_sweep_exp2}
\end{figure}

\begin{figure}[h!]
    \centering
    \includegraphics[width=0.8\columnwidth]{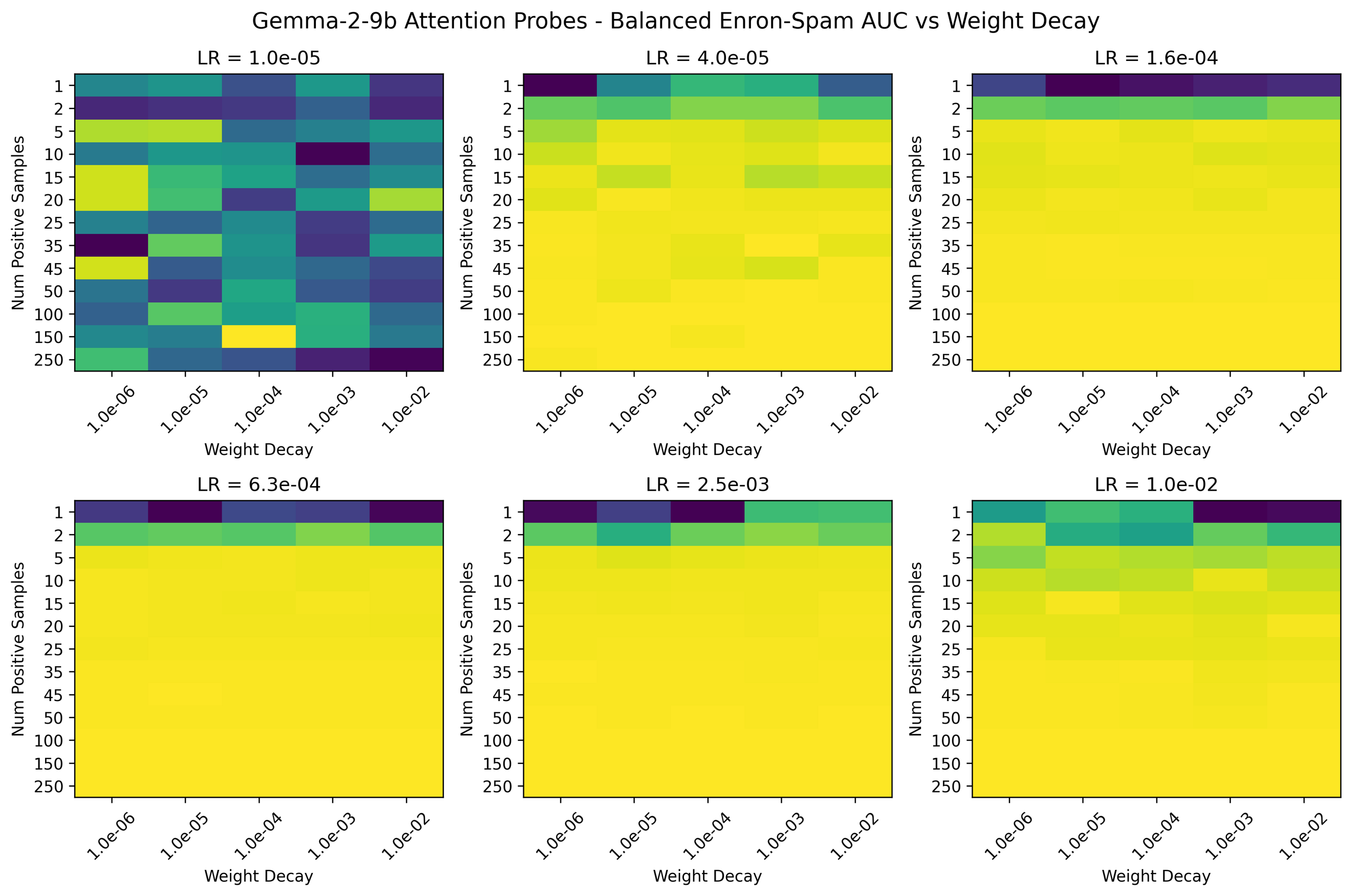}
    \caption{AUC performance for Attention probes across the weight decay sweep on the Enron-Spam dataset (Balanced training).}
    \label{fig:attention_sweep_exp4}
\end{figure}

\paragraph{Cross-Validation Results}
To simulate an expert manually selecting the best hyperparameters without sacrificing scarce training data, we present results using a 50/50 balanced cross-validation set to choose the optimal `C` value for each probe. For these plots, the number of positive samples on the x-axis refers \textit{only} to the data used to train the final probe after a hyperparameter has been selected. The data used within the cross-validation folds is treated as a separate pool for tuning and is not included in this count. The following figures compare the performance of imbalanced versus balanced training using these cross-validated probes.

\begin{figure}[h!]
    \centering
    \includegraphics[width=0.9\columnwidth]{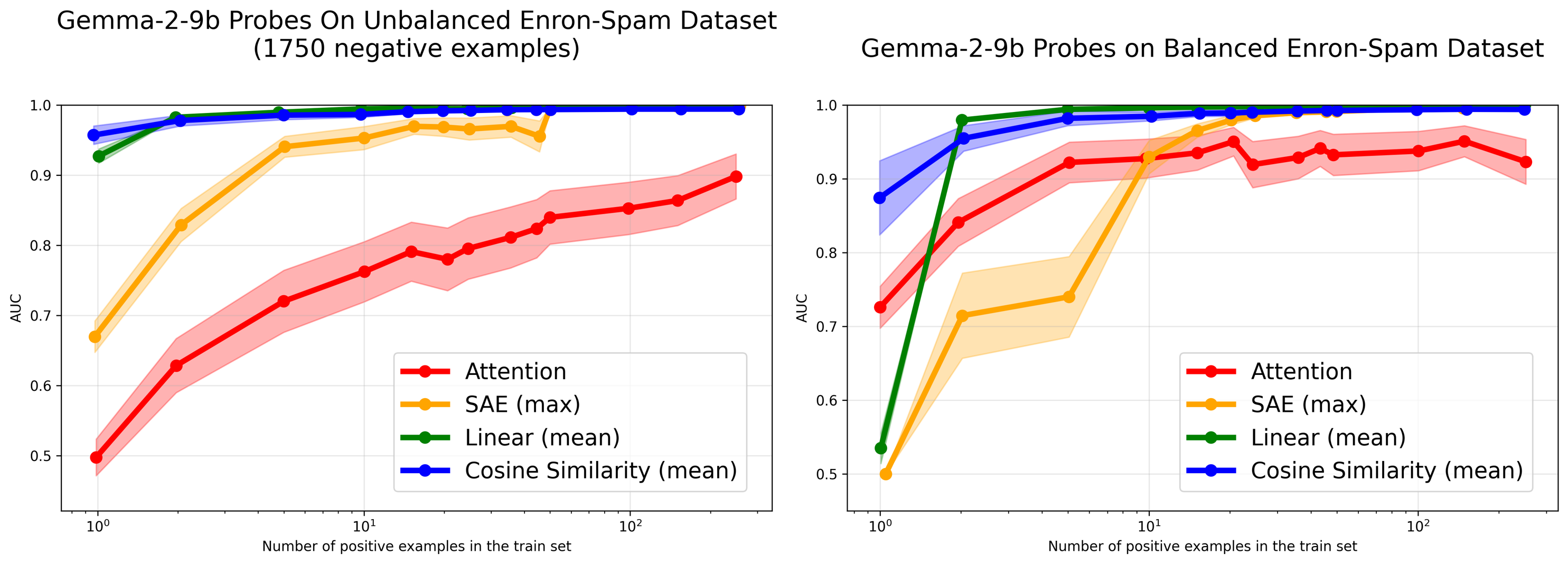}
    \caption{Cross-validated AUC performance on the Enron-Spam dataset, comparing imbalanced and balanced training.}
    \label{fig:cv_spam_auc}
\end{figure}

\begin{figure}[h!]
    \centering
    \includegraphics[width=0.9\columnwidth]{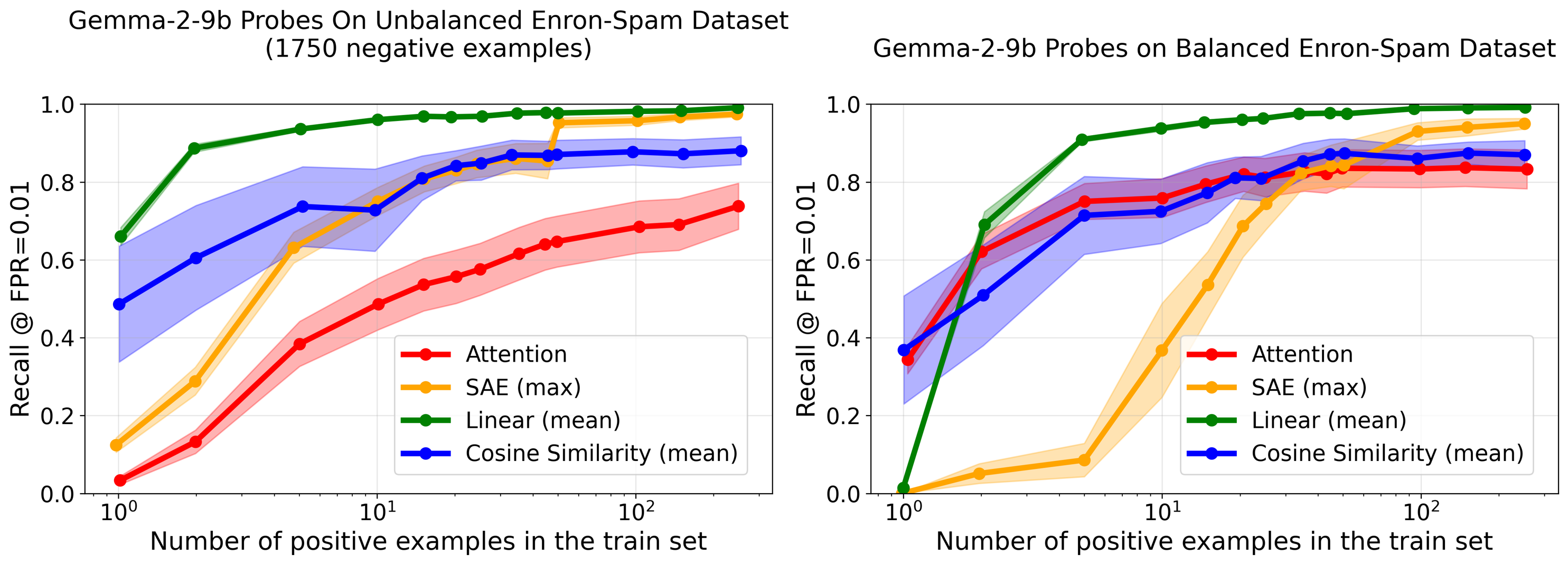}
    \caption{Cross-validated Recall @ 0.01 FPR performance on the Enron-Spam dataset.}
    \label{fig:cv_spam_recall}
\end{figure}

\begin{figure}[h!]
    \centering
    \includegraphics[width=0.9\columnwidth]{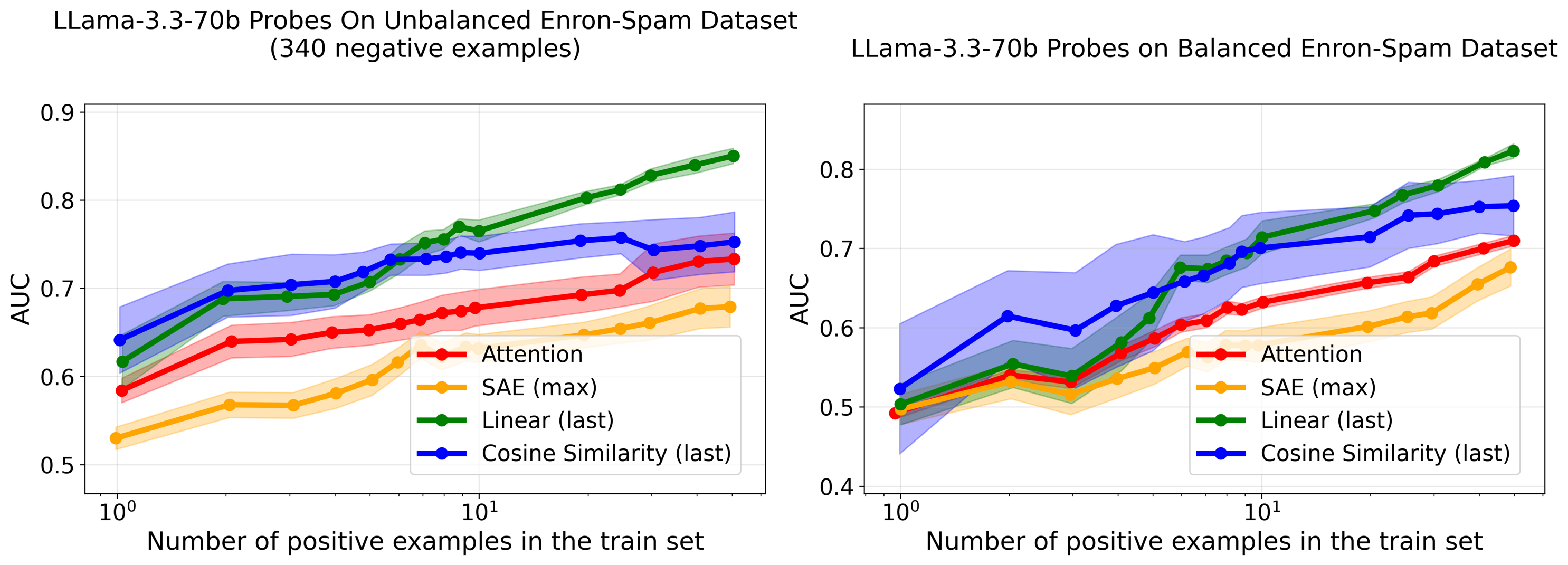}
    \caption{Cross-validated AUC performance on the MASK honesty benchmark.}
    \label{fig:cv_mask_auc}
\end{figure}

\begin{figure}[h!]
    \centering
    \includegraphics[width=0.9\columnwidth]{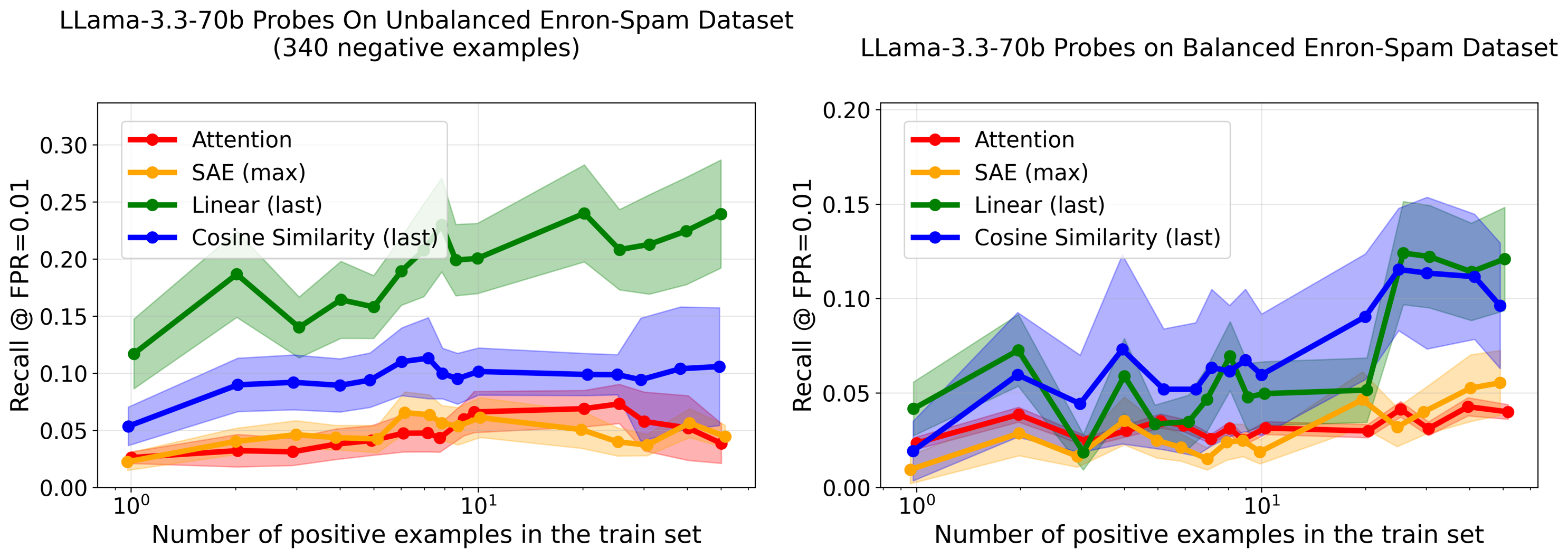}
    \caption{Cross-validated Recall @ 0.01 FPR performance on the MASK honesty benchmark.}
    \label{fig:cv_mask_recall}
\end{figure}

\subsection{Additional Imbalanced vs. Balanced Results}
\label{app:gemma_more}
This section provides additional plots comparing probe performance on imbalanced versus balanced datasets across our different evaluation settings. For high-stakes applications where monitoring is costly, Recall at a low False Positive Rate (e.g., 1\%) can be more important than AUC, as it measures the probe's ability to catch true positives while maintaining a low rate of false alarms.

\subsubsection{Gemma-2-9b Enron-Spam Results}
The following plots show detailed performance for the Enron-Spam dataset. Note that while Attention probes have lower AUC in the imbalanced setting, they often achieve high recall when trained on balanced data, suggesting they are effective at identifying positive instances when not overwhelmed by a large number of negative examples.

\begin{figure}[h!]
    \centering
    \includegraphics[width=0.7\columnwidth]{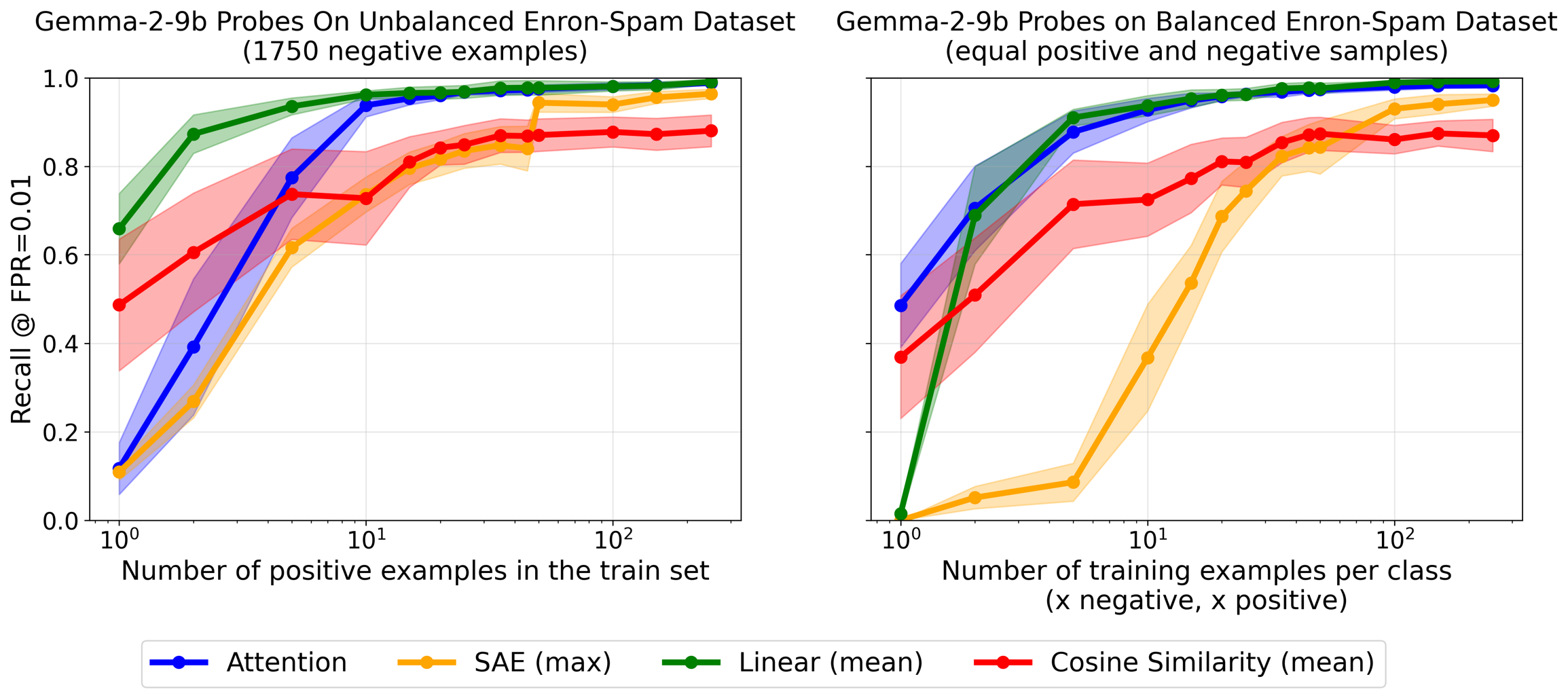}
    \caption{Recall performance of the best probes on the non-OOD spam dataset.}
    \label{fig:app_joint_recall_nood}
\end{figure}

\begin{figure}[h!]
    \centering
    \includegraphics[width=\columnwidth]{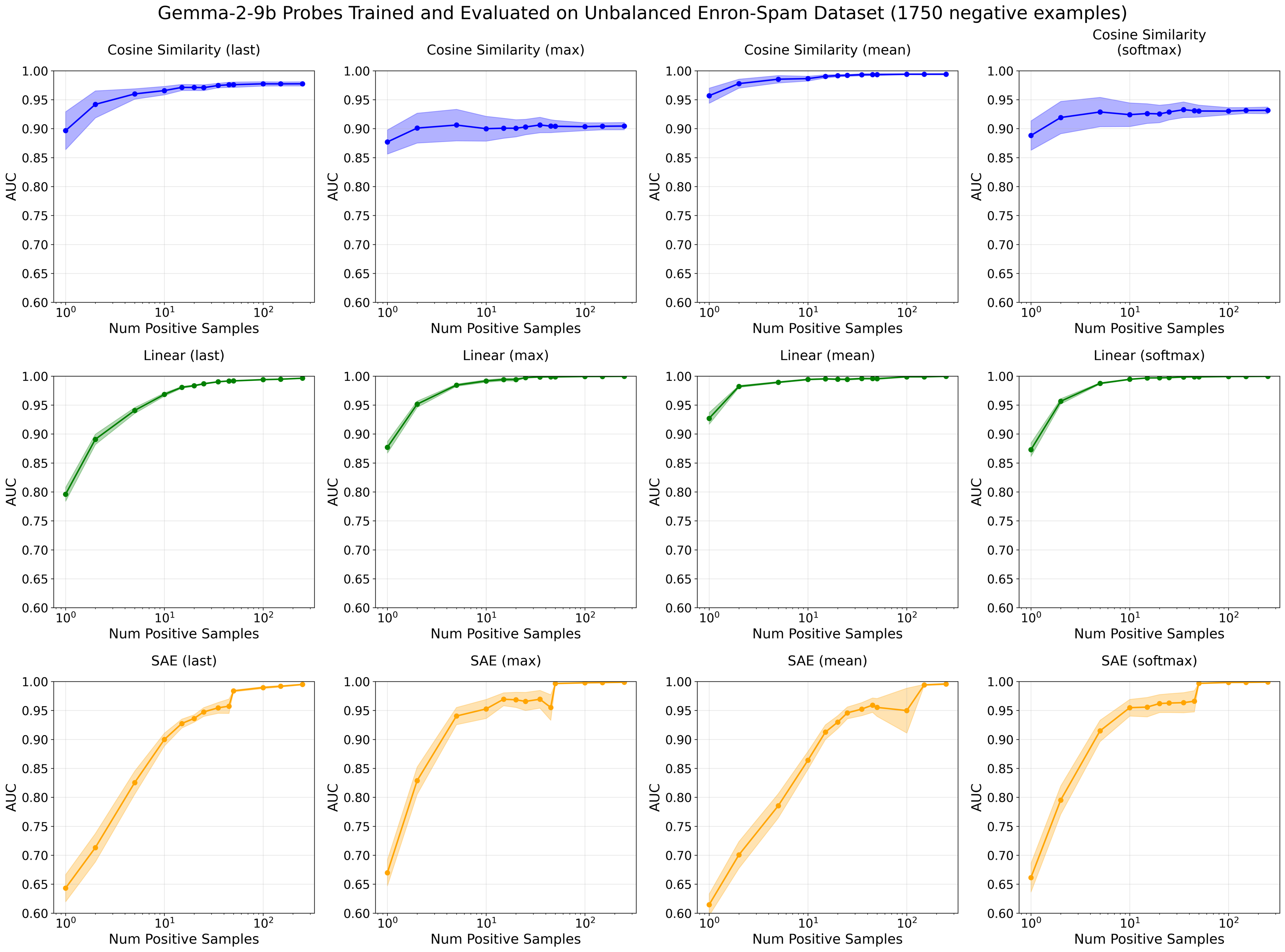}
    \caption{Per-probe AUC on the non-OOD spam dataset (Experiment 2).}
\end{figure}

\begin{figure}[h!]
    \centering
    \includegraphics[width=\columnwidth]{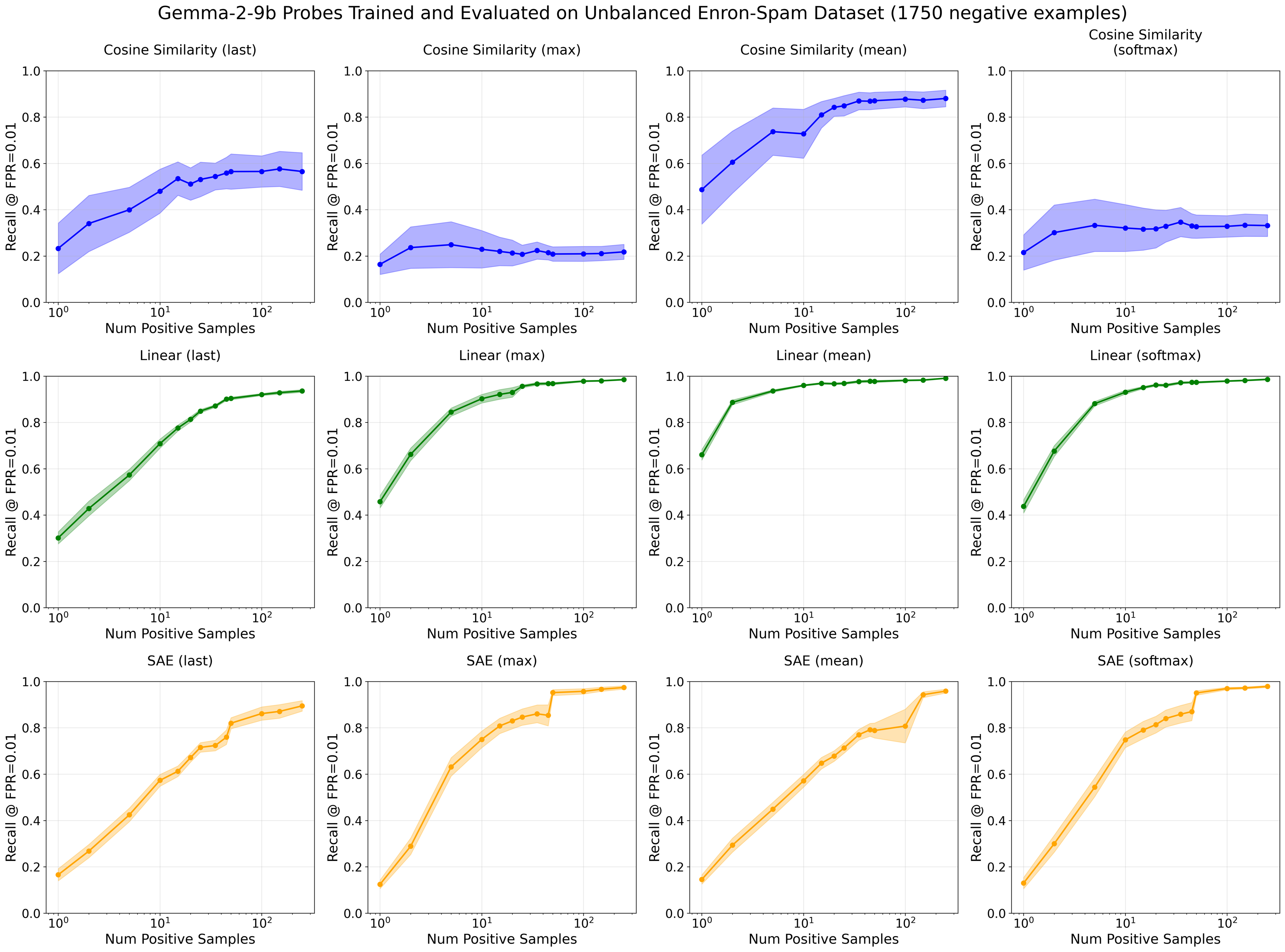}
    \caption{Per-probe Recall @ 0.01 FPR on the non-OOD spam dataset (Experiment 2).}
\end{figure}

\begin{figure}[h!]
    \centering
    \includegraphics[width=\columnwidth]{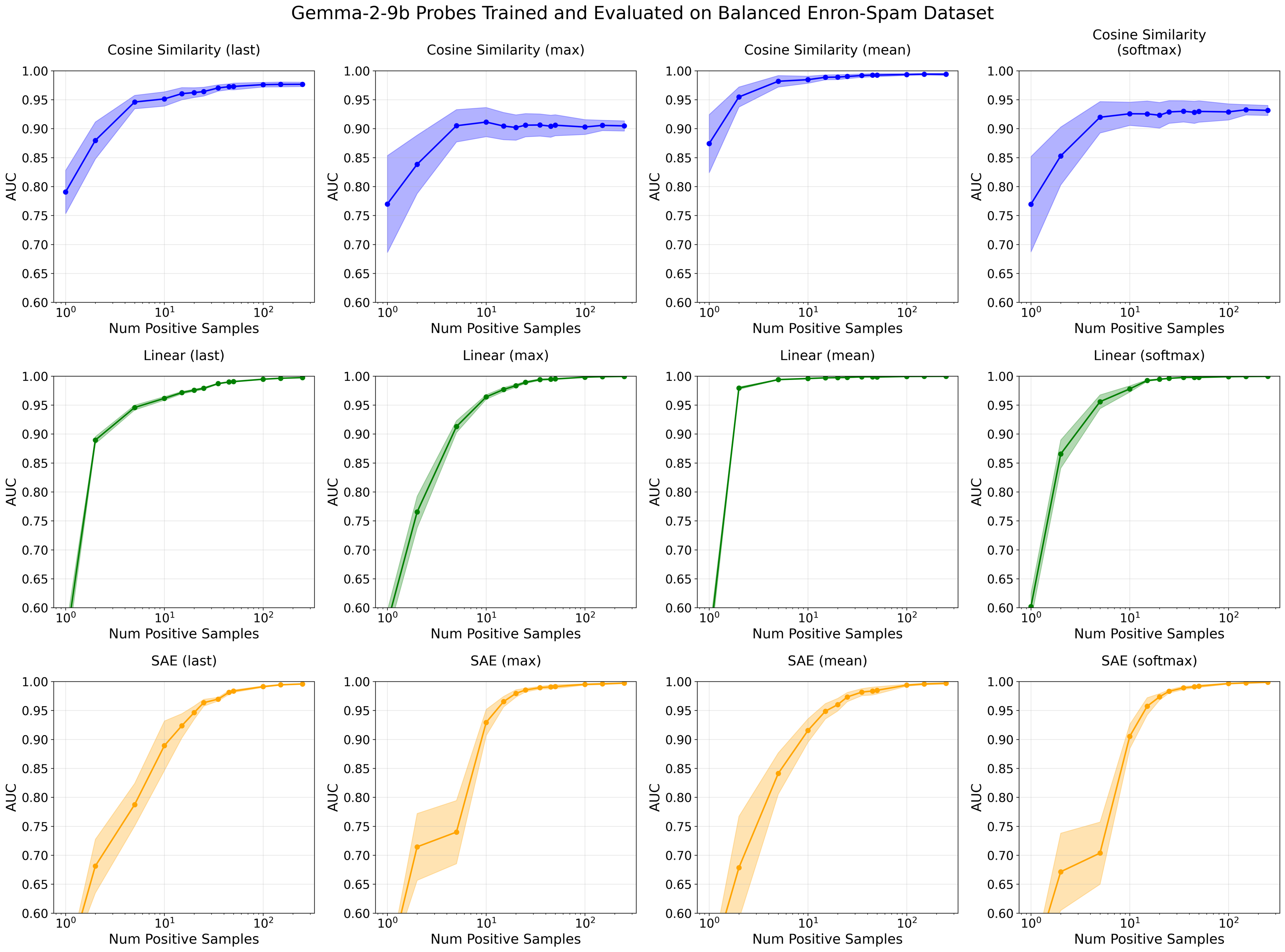}
    \caption{Per-probe AUC on the non-OOD spam dataset (Experiment 4).}
\end{figure}

\begin{figure}[h!]
    \centering
    \includegraphics[width=\columnwidth]{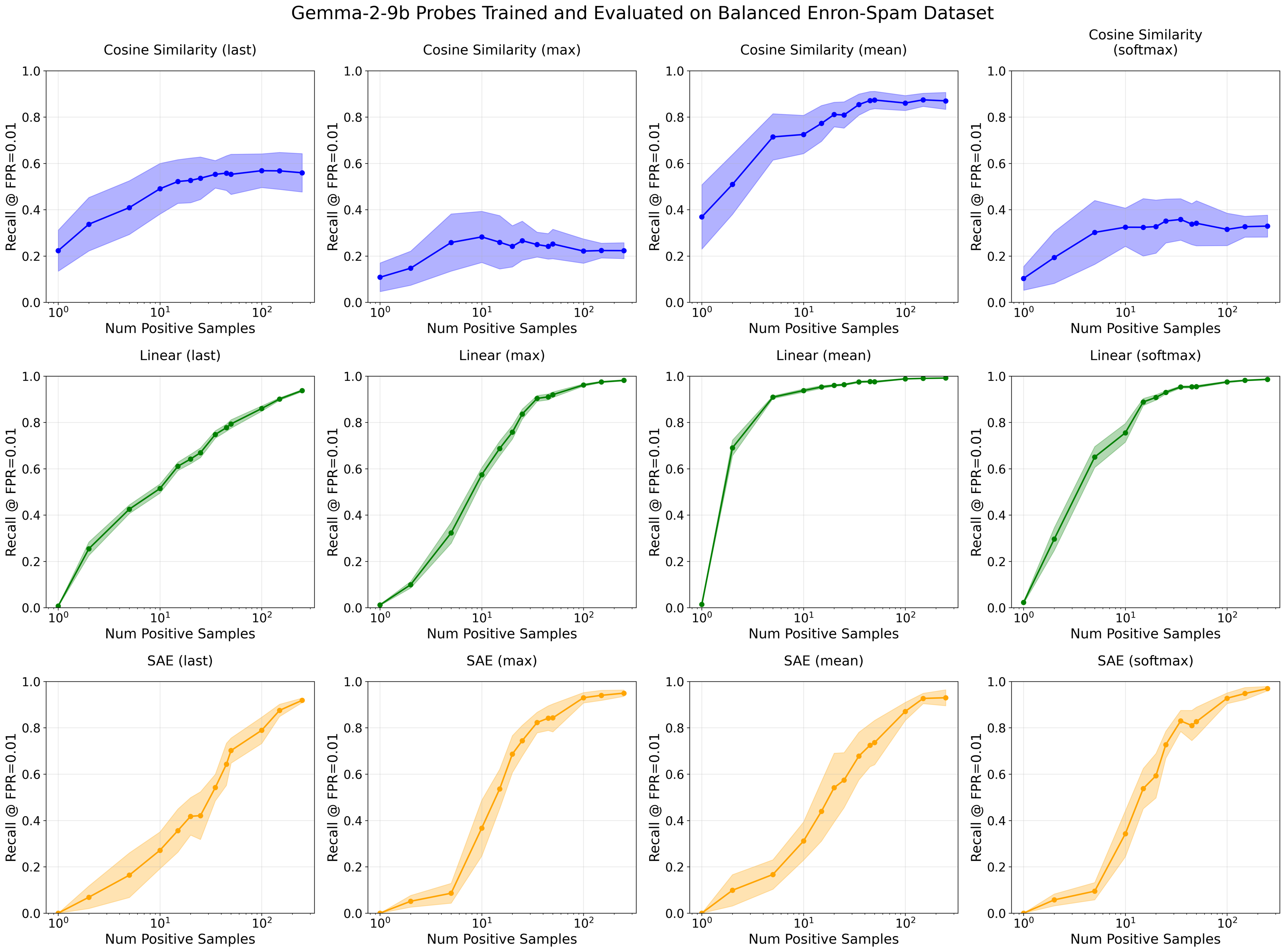}
    \caption{Per-probe Recall @ 0.01 FPR on the non-OOD spam dataset (Experiment 4).}
\end{figure}

\subsubsection{Gemma-2-9b Out-Of-Distribution Results On SMS Spam}

We test whether the benefits of imbalanced training generalize to an out-of-distribution dataset, SMS Spam \cite{team2017spam}. The results are inconclusive and suggest that the advantages do not clearly transfer. For the top 5 performing probes in the 1-10 positive sample range, imbalanced training yields a mixed result of \textbf{+0.012 AUC} but \textbf{-0.031 Recall} points. In higher data regimes (50-150 samples), the performance change is more clearly negative, at \textbf{-0.011 AUC} and \textbf{-0.059 Recall} points. This indicates that while imbalanced training is highly effective for in-distribution data, its benefits for out-of-distribution generalization are not guaranteed.

\begin{figure}[h!]
    \centering
    \includegraphics[width=0.8\columnwidth]{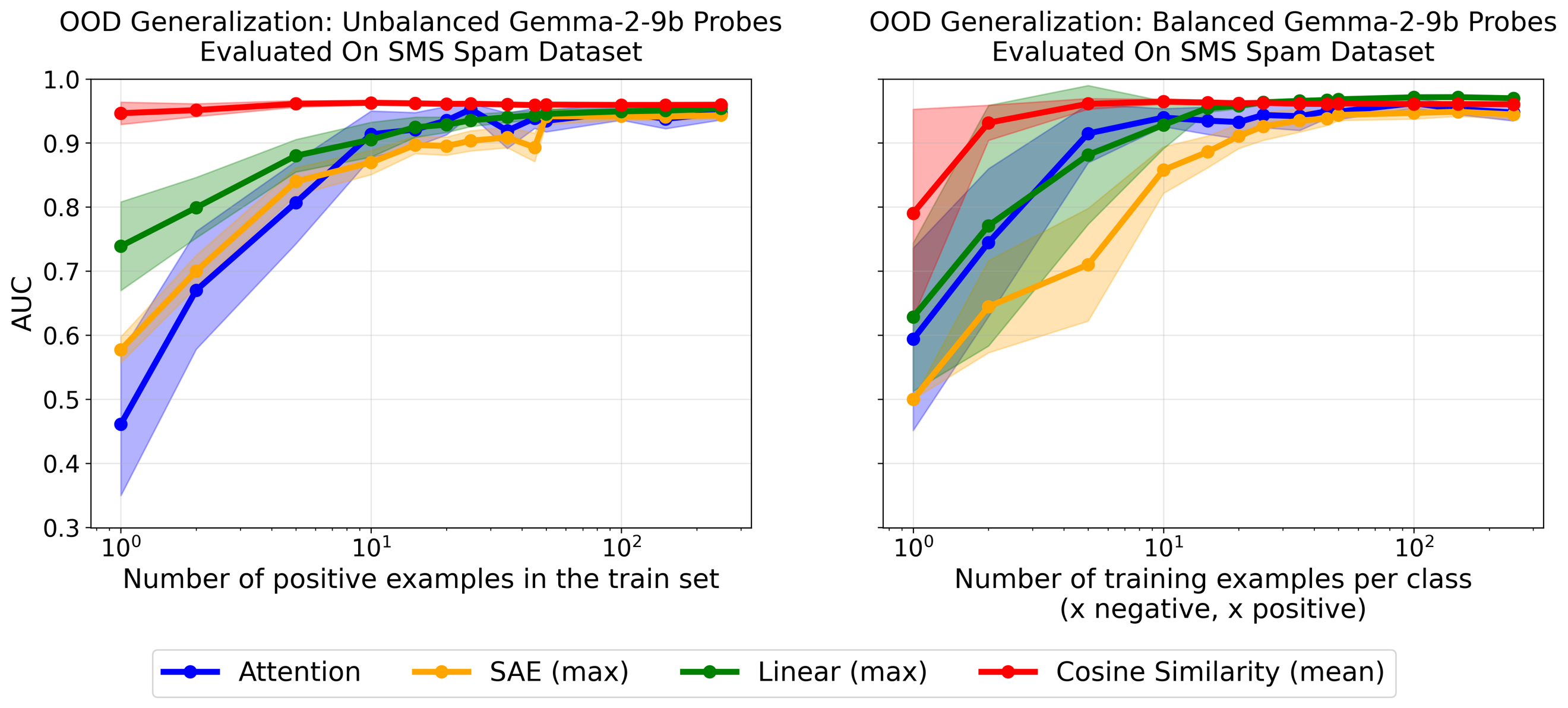}
    \caption{AUC performance of best probes on the SMS spam dataset.}
\end{figure}

\begin{figure}[h!]
    \centering
    \includegraphics[width=0.8\columnwidth]{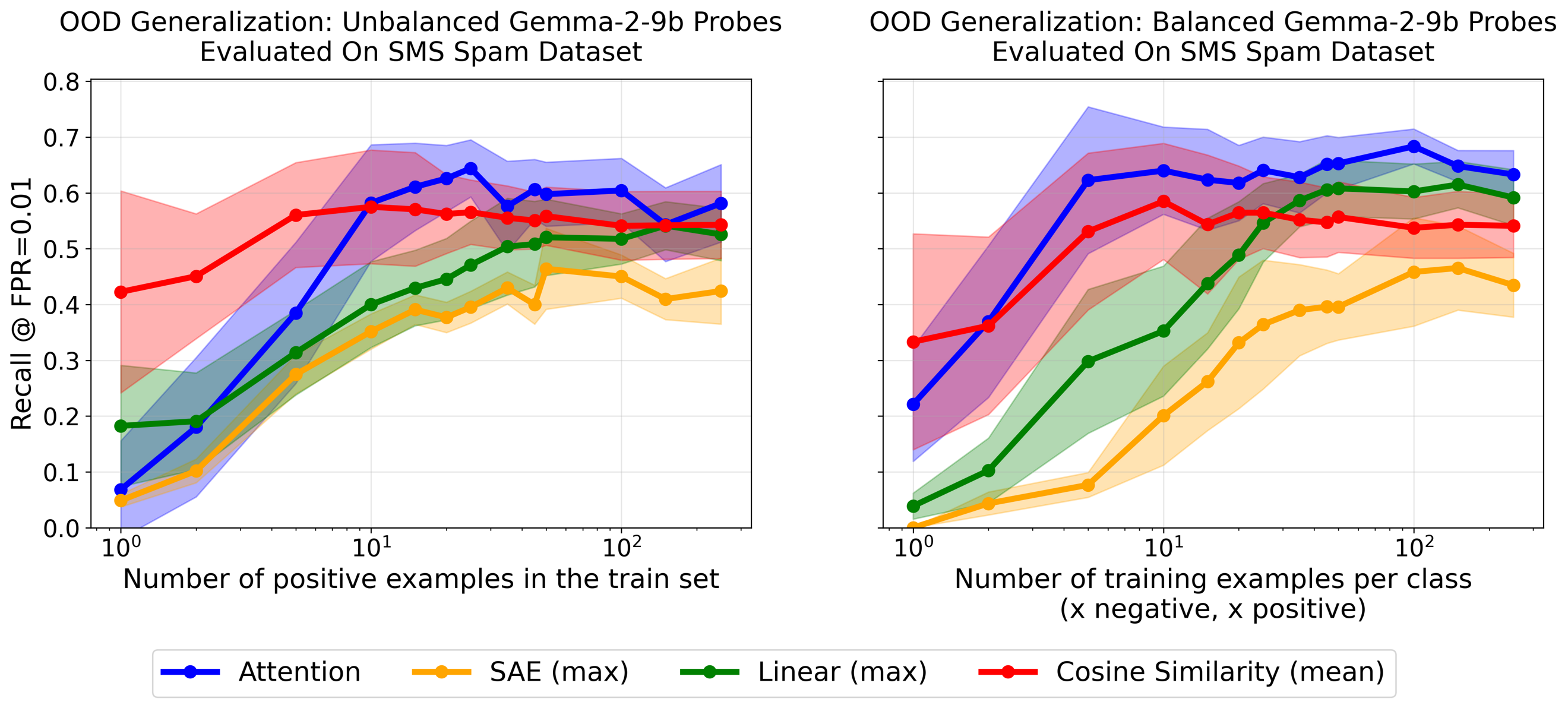}
    \caption{Recall @ 0.01 FPR of best probes on the SMS spam dataset.}
\end{figure}

\begin{figure}[h!]
    \centering
    \includegraphics[width=\columnwidth]{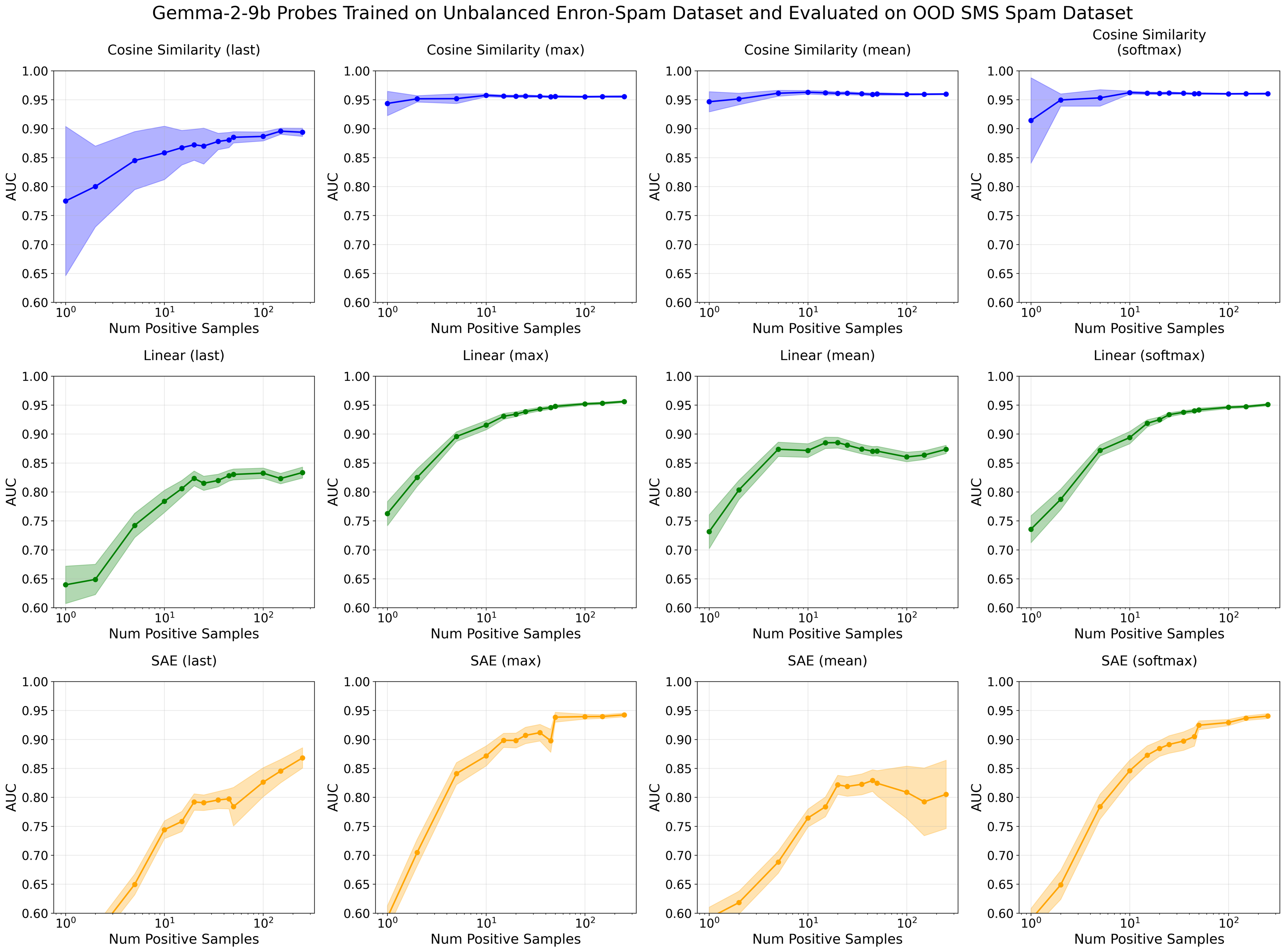}
    \caption{Per-probe AUC on the SMS spam generalization task.}
\end{figure}

\begin{figure}[h!]
    \centering
    \includegraphics[width=\columnwidth]{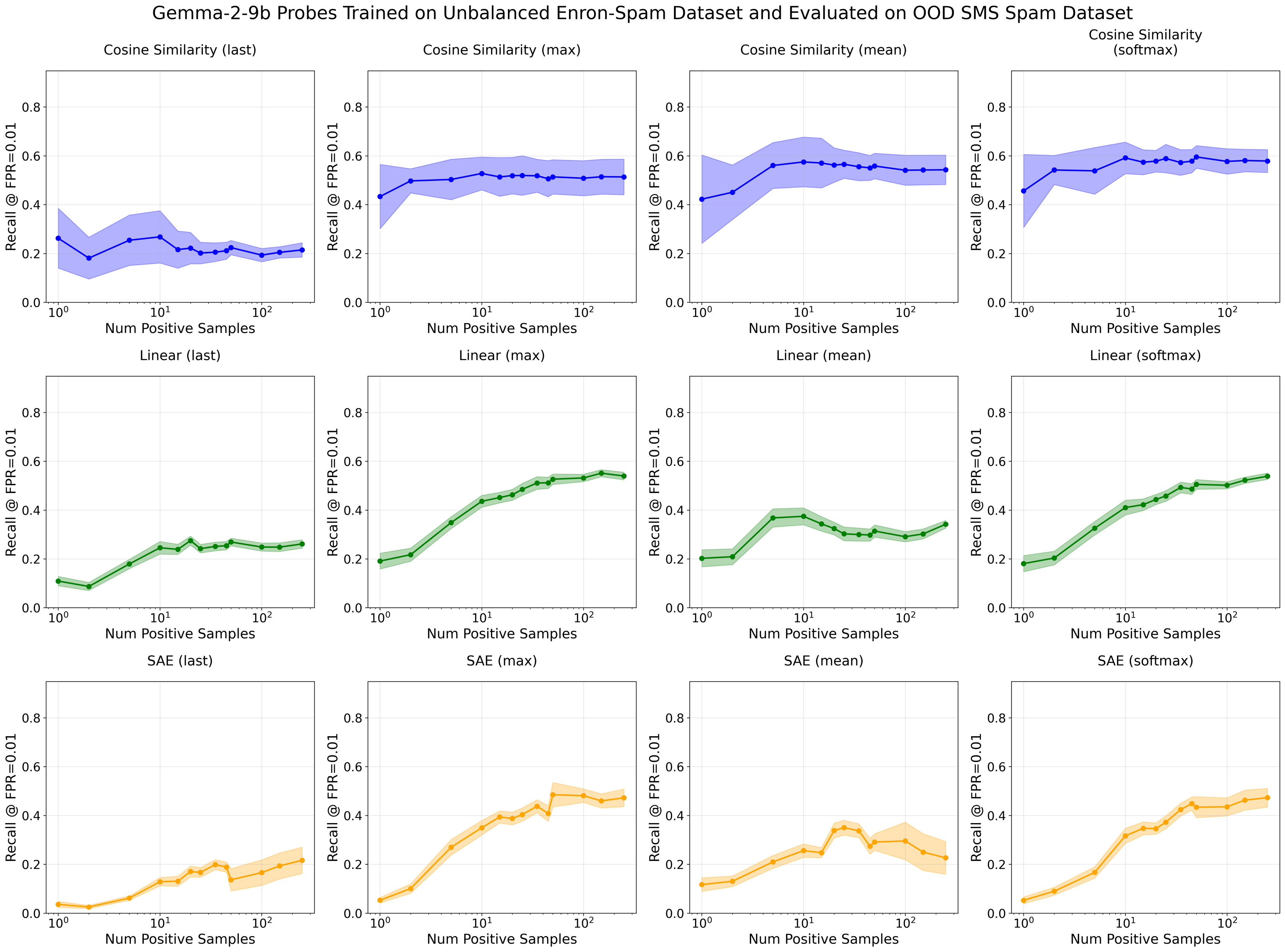}
    \caption{Per-probe Recall @ 0.01 FPR on the SMS spam generalization task.}
\end{figure}

\subsubsection{Llama-3.3-70B-Instruct MASK Benchmark Results}
\begin{figure}[h!]
    \centering
    \includegraphics[width=0.7\columnwidth]{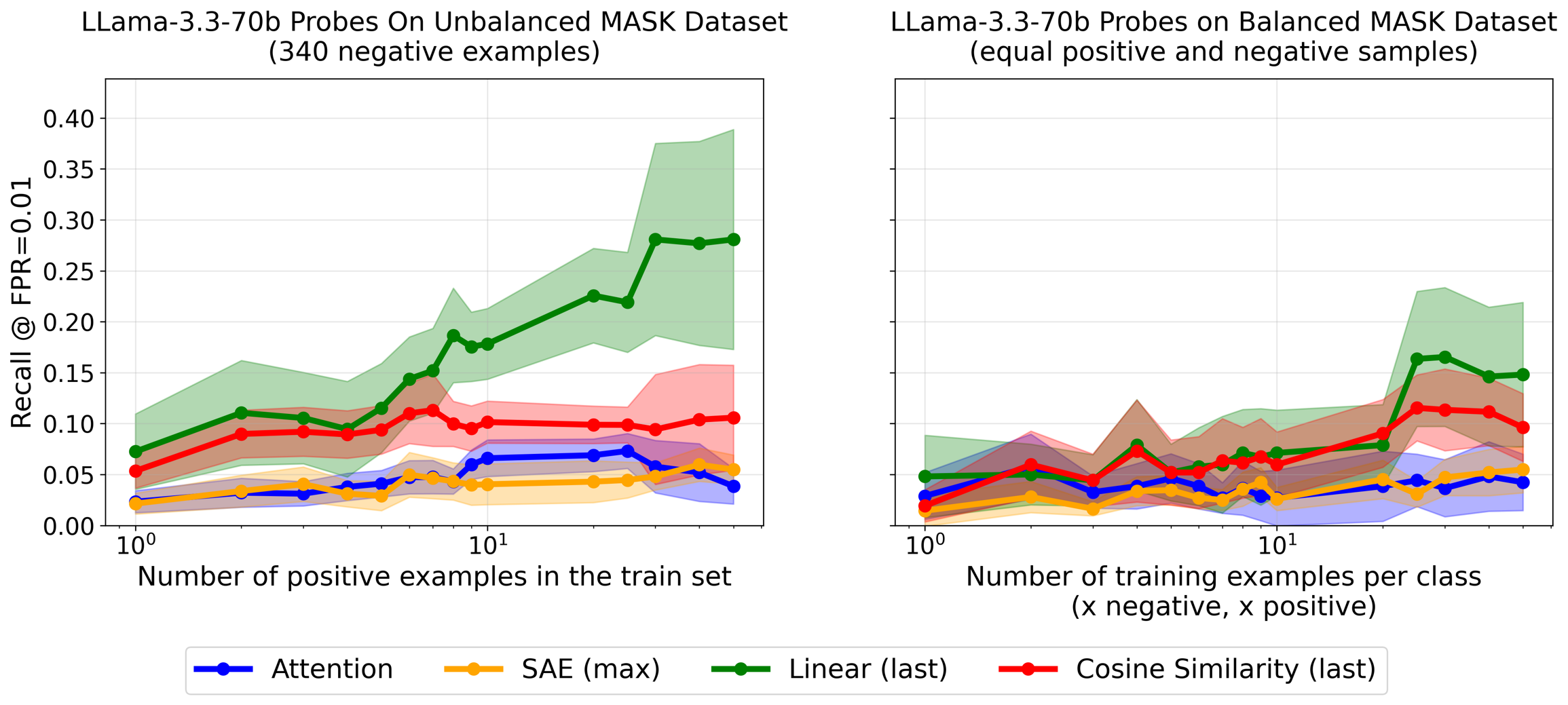}
    \caption{Recall performance of the best probes on the MASK honesty dataset.}
\end{figure}

\begin{figure}[h!]
    \centering
    \includegraphics[width=\columnwidth]{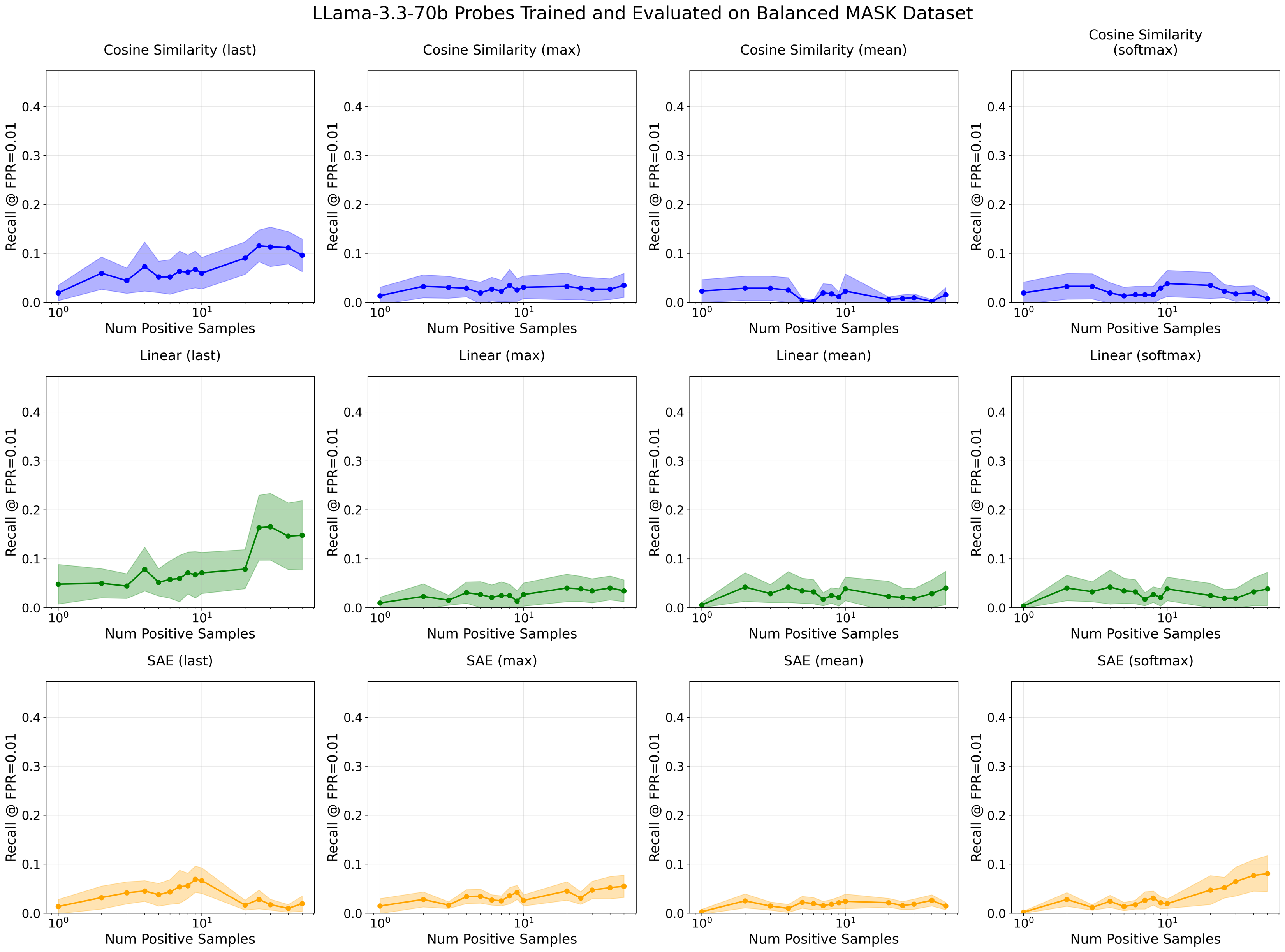}
    \caption{Detailed per-probe recall performance for Experiment 4 on the MASK dataset.}
\end{figure}

\subsection{Additional LLM Upsampling Results}
\label{app:upsampling_plots}
This section provides detailed plots for our LLM upsampling experiments.

\subsubsection{Gemma-2-9b Enron-Spam Results}
We observe varied effects across probe types. Interestingly, for SAE probes, performance continues to improve even at high upsampling ratios (10x, 20x). In contrast, Cosine Similarity probes degrade significantly with more upsampling. This may be because their non-parametric nature makes them highly sensitive to the training data distribution; low-diversity synthetic samples could shift the class centroid in a non-robust way. Linear probes show moderate benefits, with performance tapering off at the highest ratios.

\begin{figure}[h!]
    \centering
    \includegraphics[width=\columnwidth]{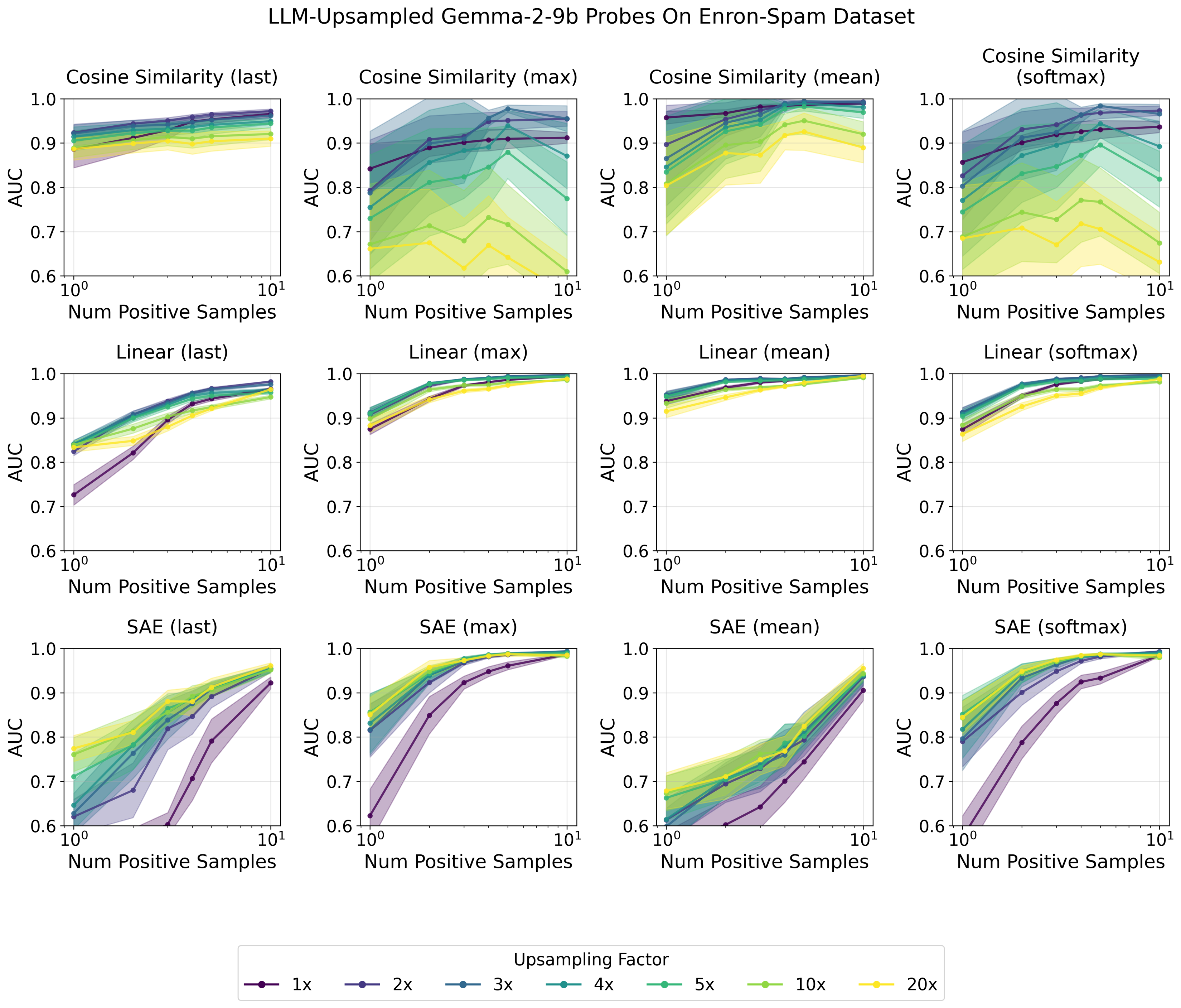}
    \caption{Per-probe AUC with LLM upsampling on the Enron-Spam dataset.}
\end{figure}

\begin{figure}[h!]
    \centering
    \includegraphics[width=\columnwidth]{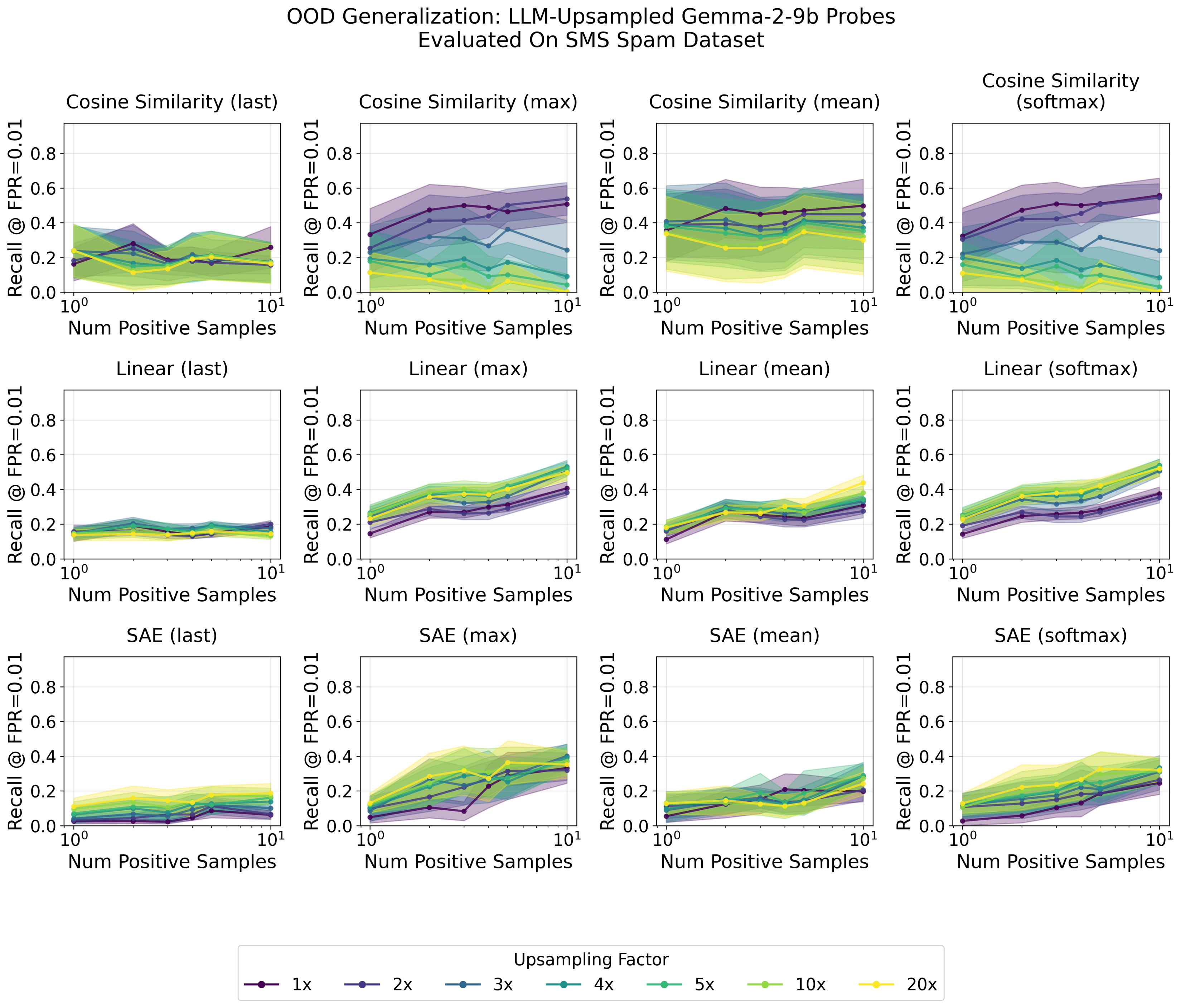}
    \caption{Per-probe Recall @ 0.01 FPR with LLM upsampling on the Enron-Spam dataset.}
\end{figure}

\begin{figure}[h!]
    \centering
    \begin{subfigure}{0.49\columnwidth}
        \includegraphics[width=\linewidth]{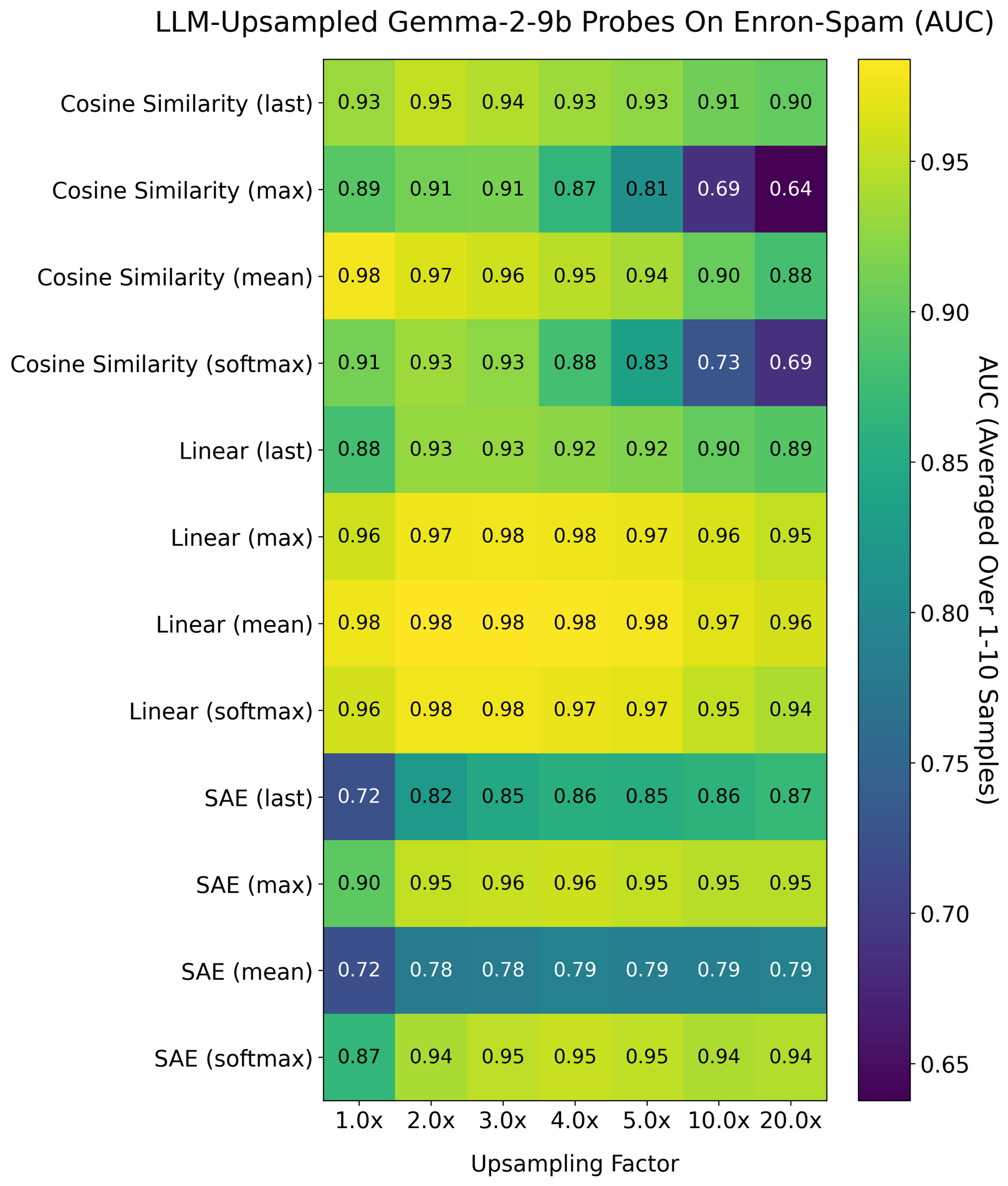}
        \caption{AUC Heatmap}
    \end{subfigure}
    \begin{subfigure}{0.49\columnwidth}
        \includegraphics[width=\linewidth]{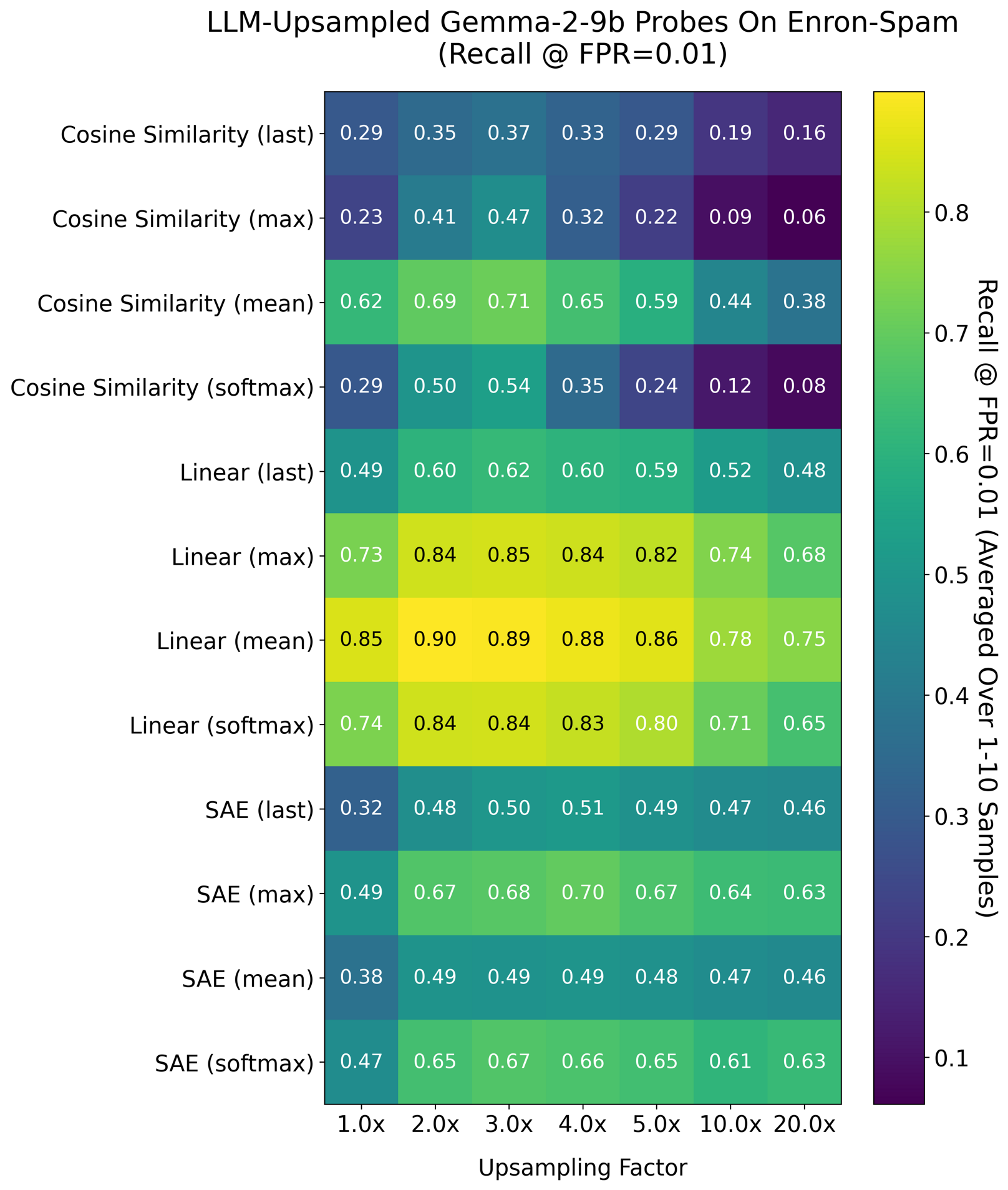}
        \caption{Recall @ 0.01 FPR Heatmap}
    \end{subfigure}
    \caption{Heatmaps showing performance across upsampling factors and number of real positive examples on the Enron-Spam dataset.}
\end{figure}

\subsubsection{Generalization Results on SMS Spam}
The generalization results show a similar pattern: moderate upsampling (3x-5x) tends to improve performance on the OOD dataset, but higher ratios do not consistently help and can degrade performance.

\begin{figure}[h!]
    \centering
    \begin{subfigure}{0.49\columnwidth}
        \includegraphics[width=\linewidth]{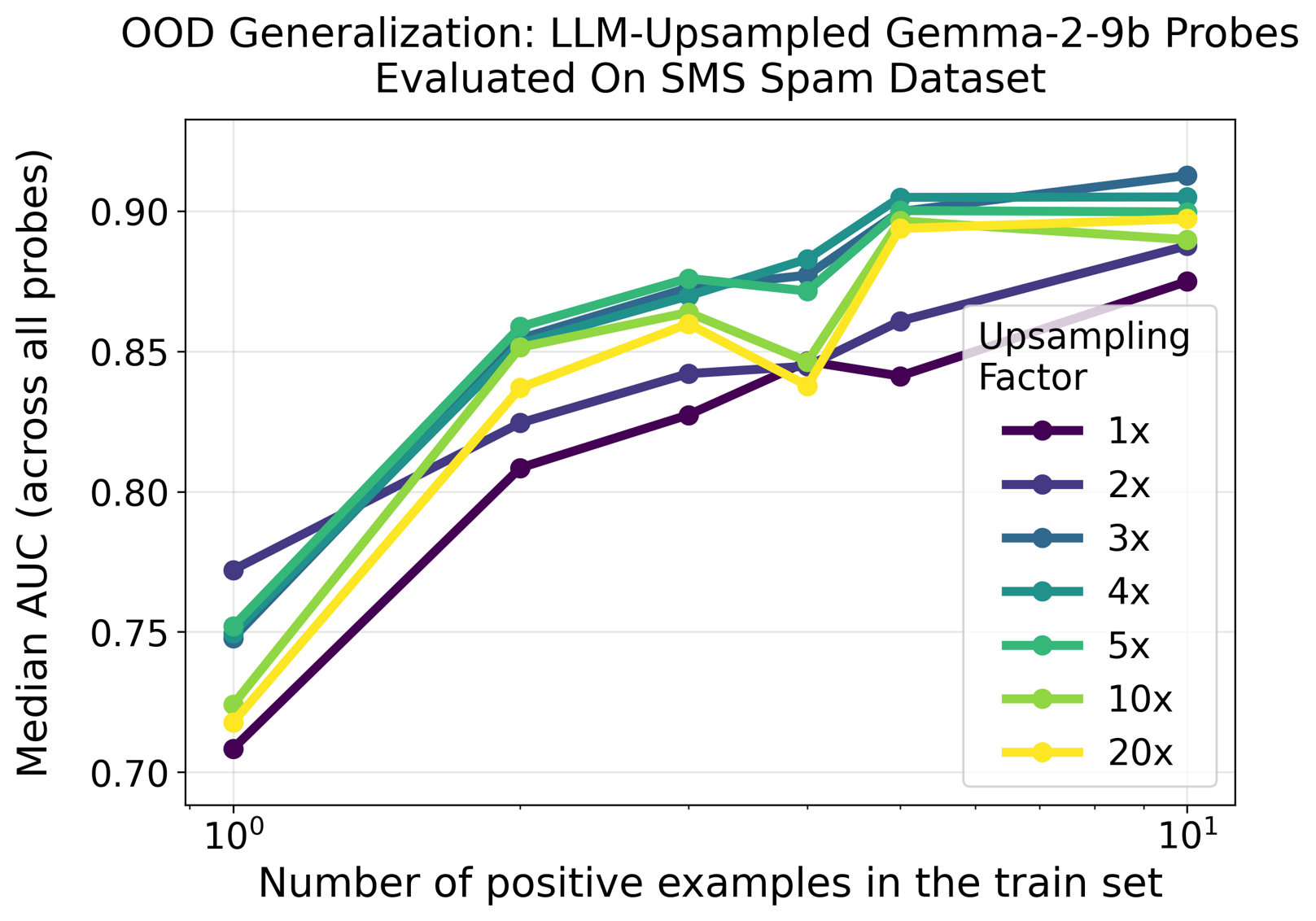}
        \caption{AUC}
        \label{fig:upsampling_gen_auc}
    \end{subfigure}
    \hfill
    \begin{subfigure}{0.49\columnwidth}
        \includegraphics[width=\linewidth]{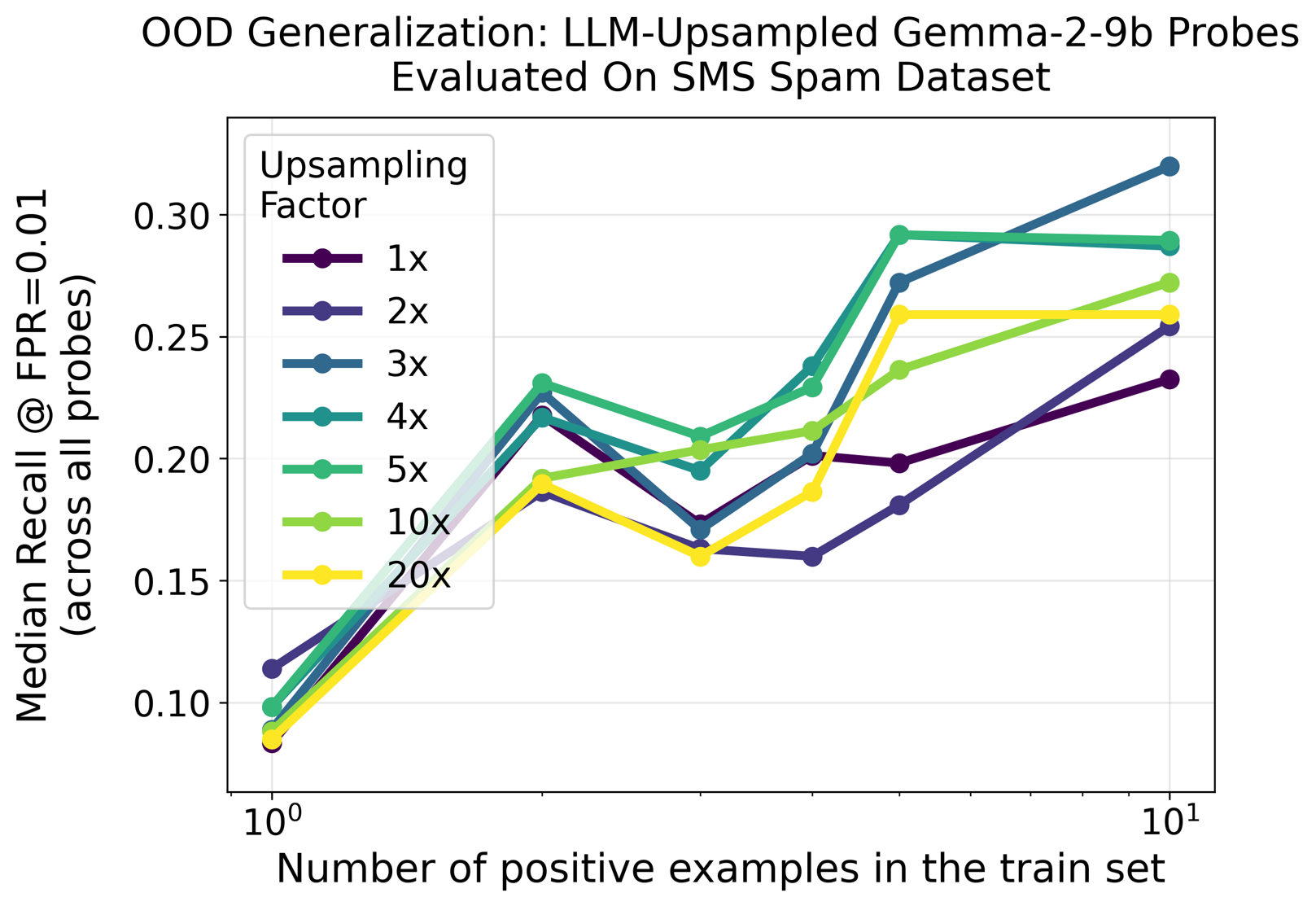}
        \caption{Recall @ 0.01 FPR}
        \label{fig:upsampling_gen_recall}
    \end{subfigure}
    \caption{Aggregated performance with LLM upsampling on the generalization SMS Spam dataset.}
    \label{fig:upsampling_gen_joint}
\end{figure}

\begin{figure}[h!]
    \centering
    \begin{subfigure}{0.49\columnwidth}
        \includegraphics[width=\linewidth]{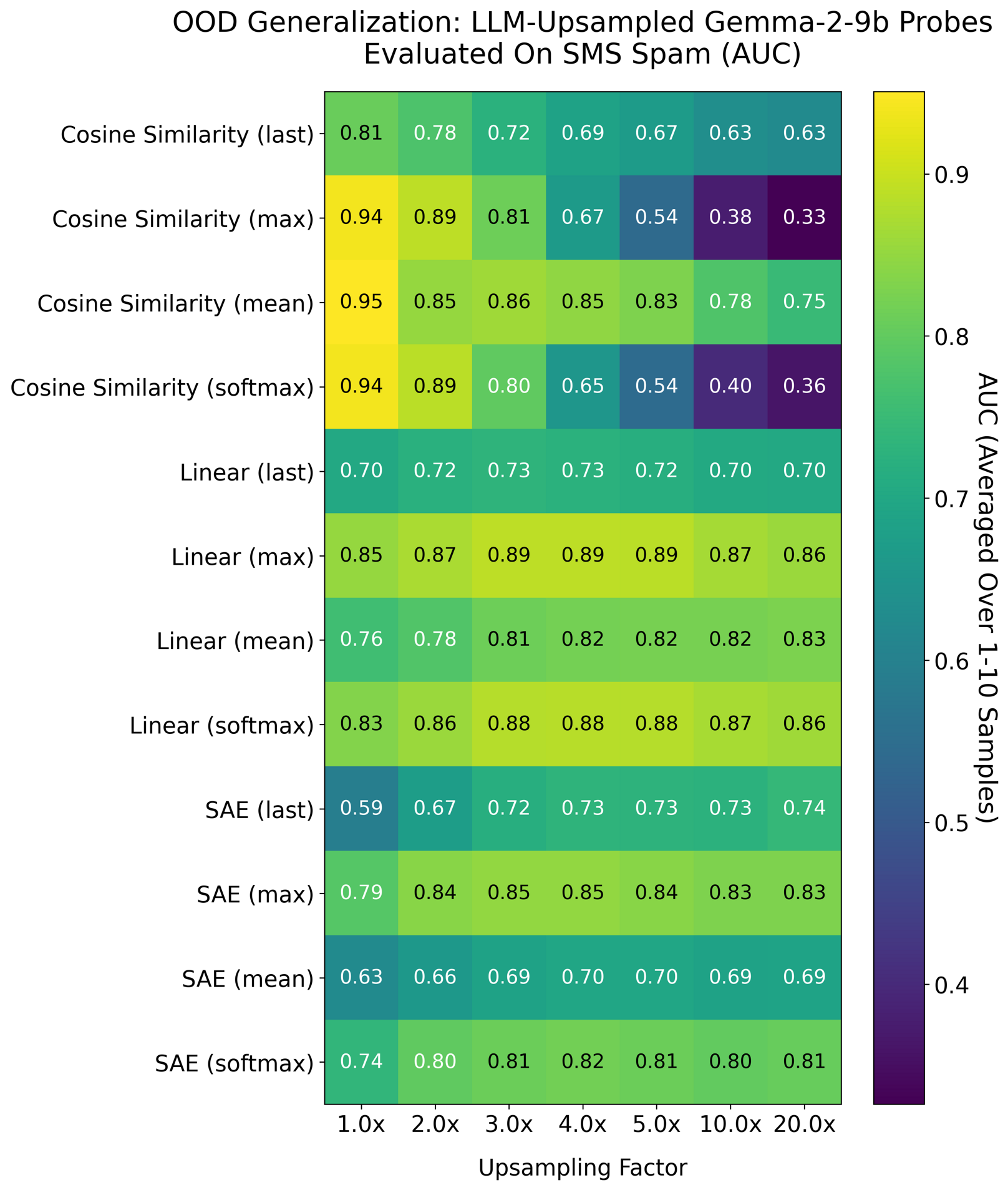}
        \caption{AUC Heatmap}
    \end{subfigure}
    \hfill
    \begin{subfigure}{0.49\columnwidth}
        \includegraphics[width=\linewidth]{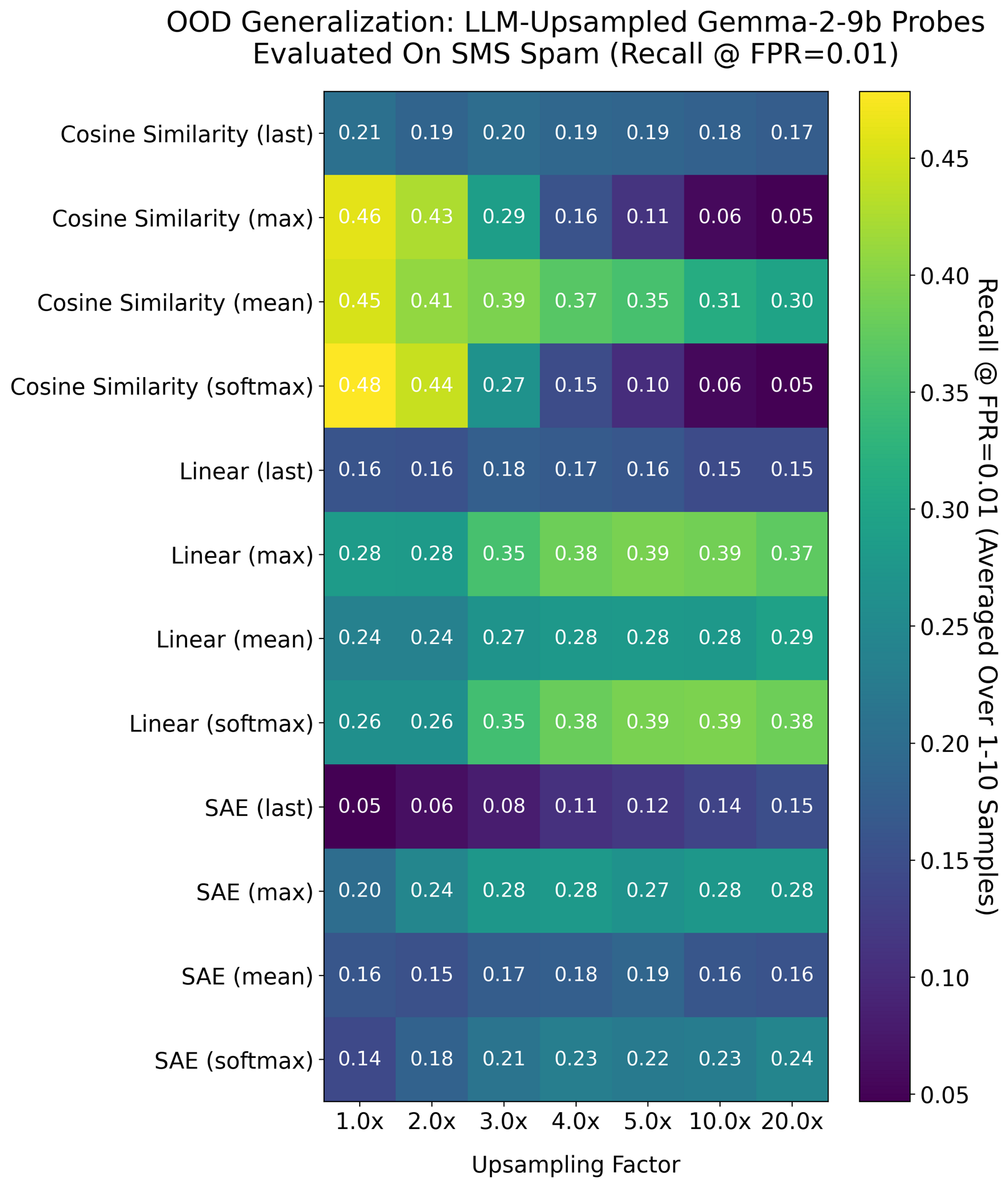}
        \caption{Recall @ 0.01 FPR Heatmap}
    \end{subfigure}
    \caption{Heatmaps showing generalization performance across upsampling factors and number of real positive examples.}
\end{figure}

\subsection{Additional Scaling Law Results}
\label{app:scaling_plots}
This section provides detailed plots for our scaling experiments across the Qwen-3 model family.

\subsubsection{Enron-Spam Results}
\begin{figure}[h!]
    \centering
    \begin{subfigure}{0.49\columnwidth}
        \includegraphics[width=\linewidth]{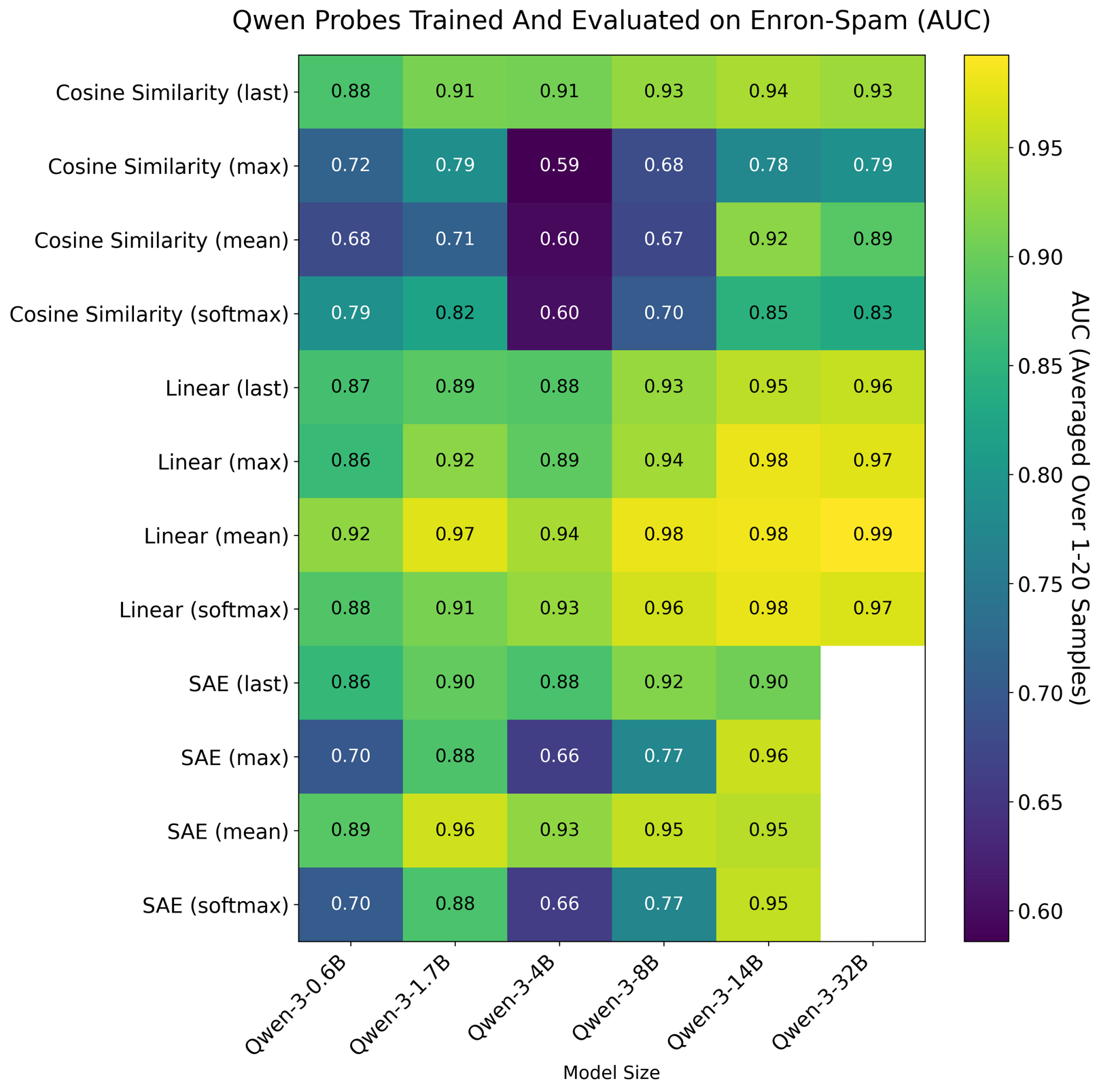}
        \caption{AUC Heatmap}
    \end{subfigure}
    \hfill
    \begin{subfigure}{0.49\columnwidth}
        \includegraphics[width=\linewidth]{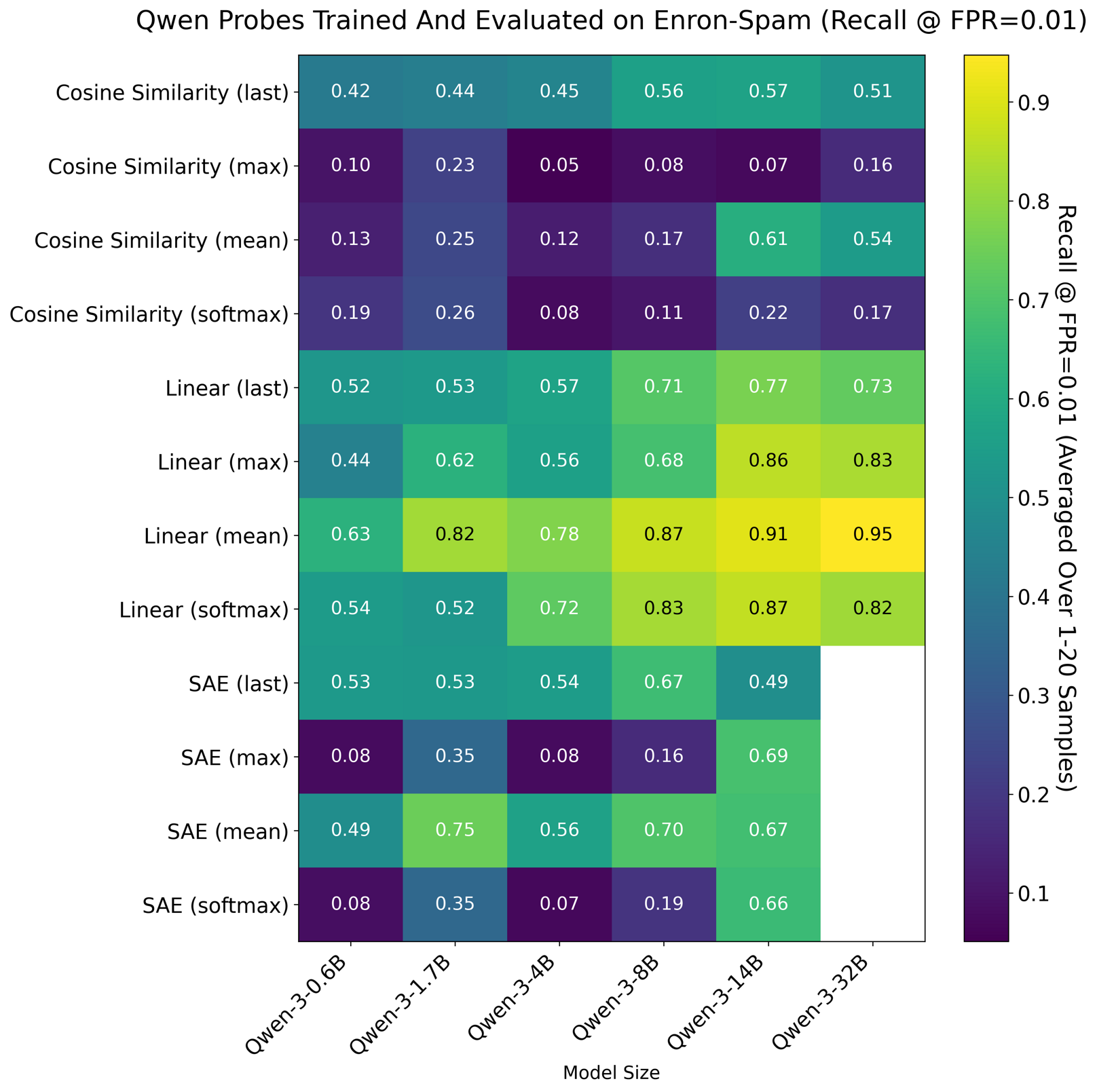}
        \caption{Recall @ 0.01 FPR Heatmap}
    \end{subfigure}
    \caption{Heatmaps showing performance across model sizes and number of positive examples. Note that results for Qwen-32B are omitted for SAE probes as no open-source transcoders/SAEs were available.}
\end{figure}

\subsubsection{Generalization Results on SMS Spam}
The scaling trend largely holds for generalization, where larger models tend to perform better. However, the results are not monotonic, with some smaller models (e.g., Qwen-3-1.7B) outperforming larger ones (Qwen-3-4B), suggesting that factors beyond just scale can influence the ease of probing a model's internal representations.

\begin{figure}[h!]
    \centering
    \begin{subfigure}{0.49\columnwidth}
        \includegraphics[width=\linewidth]{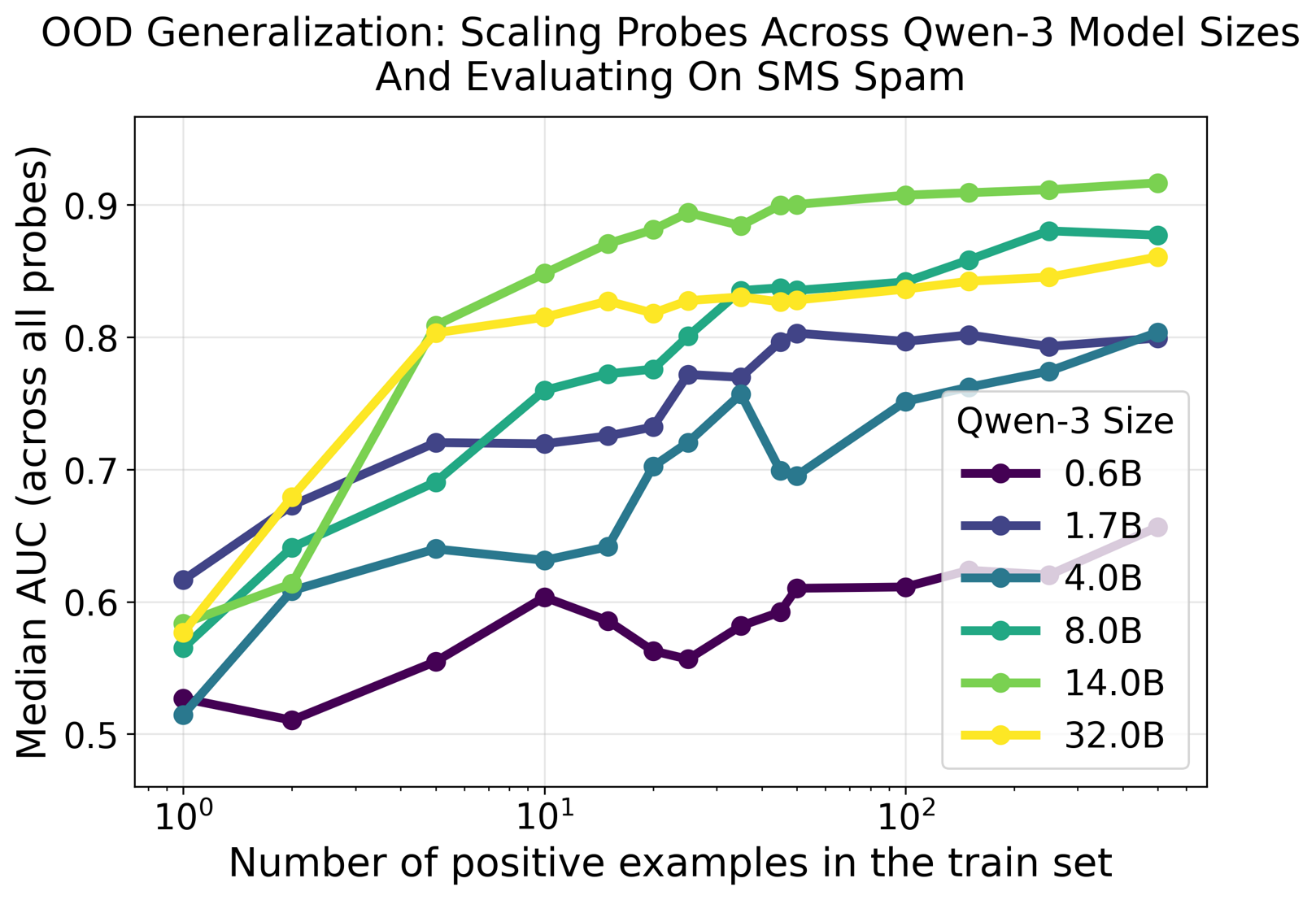}
        \caption{AUC}
        \label{fig:scaling_gen_auc}
    \end{subfigure}
    \hfill
    \begin{subfigure}{0.49\columnwidth}
        \includegraphics[width=\linewidth]{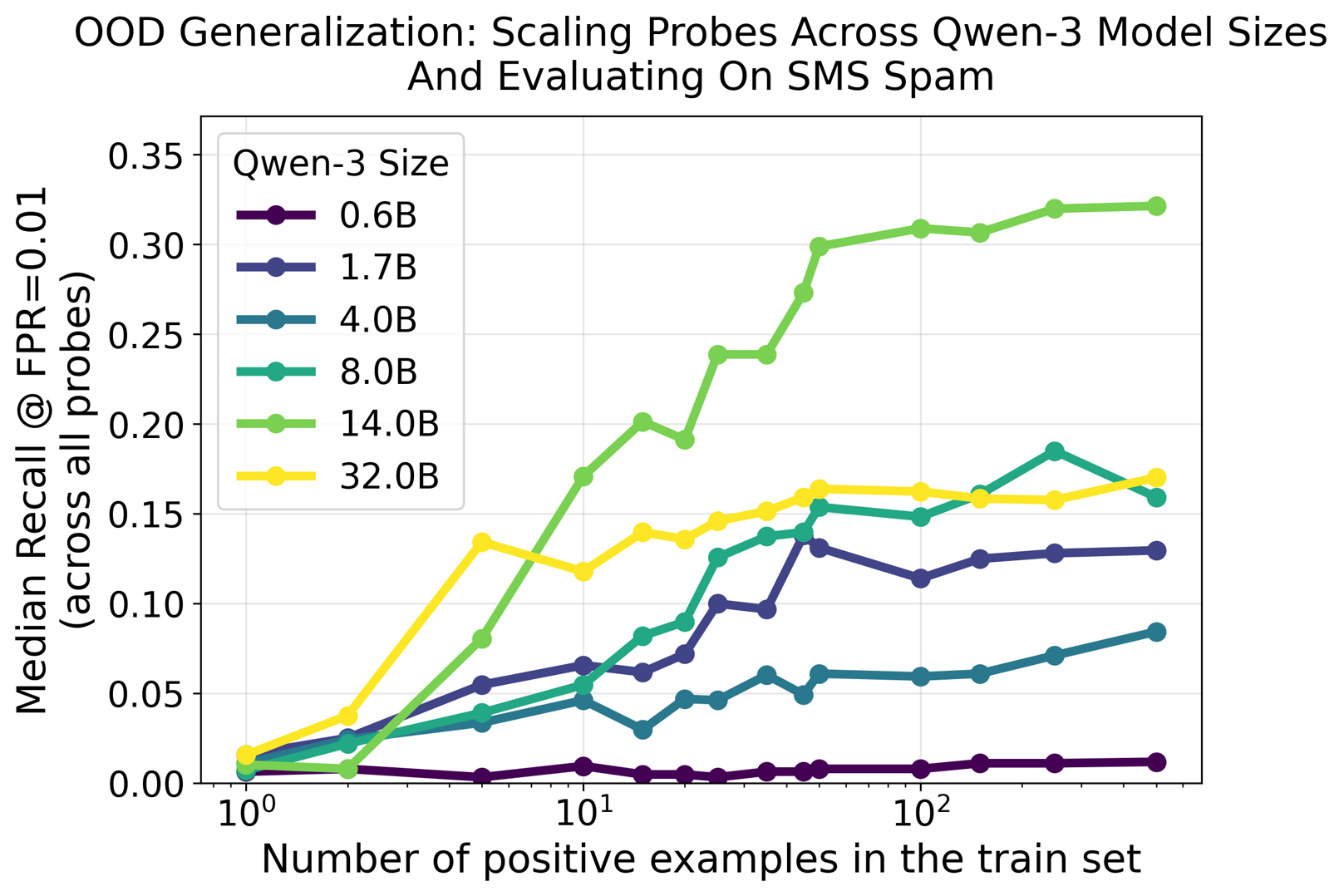}
        \caption{Recall @ 0.01 FPR}
        \label{fig:scaling_gen_recall}
    \end{subfigure}
    \caption{Aggregated performance across Qwen-3 model sizes on the generalization SMS Spam dataset.}
    \label{fig:scaling_gen_joint}
\end{figure}

\begin{figure}[h!]
    \centering
    \begin{subfigure}{0.49\columnwidth}
        \includegraphics[width=\linewidth]{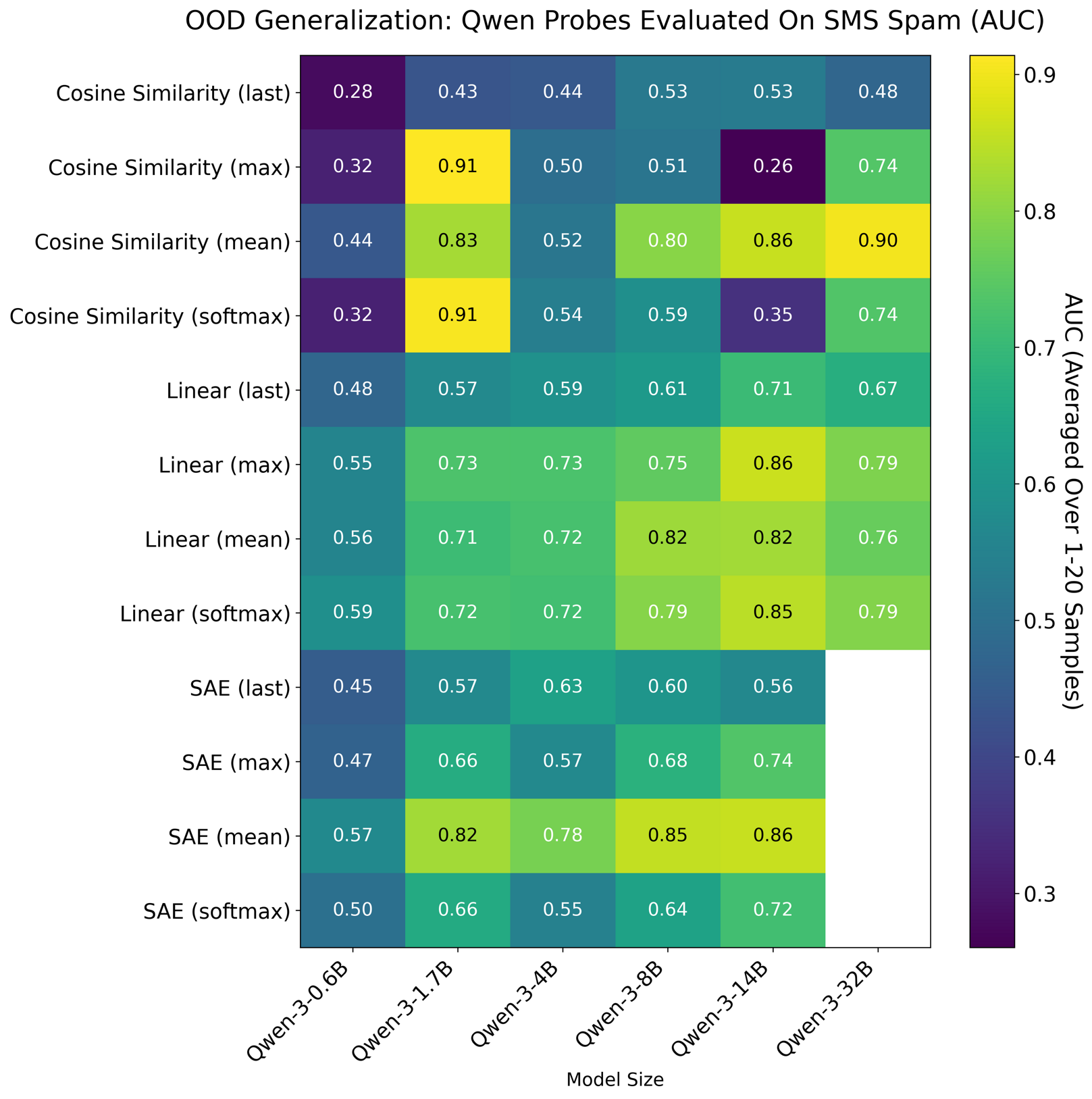}
        \caption{AUC Heatmap}
    \end{subfigure}
    \hfill
    \begin{subfigure}{0.49\columnwidth}
        \includegraphics[width=\linewidth]{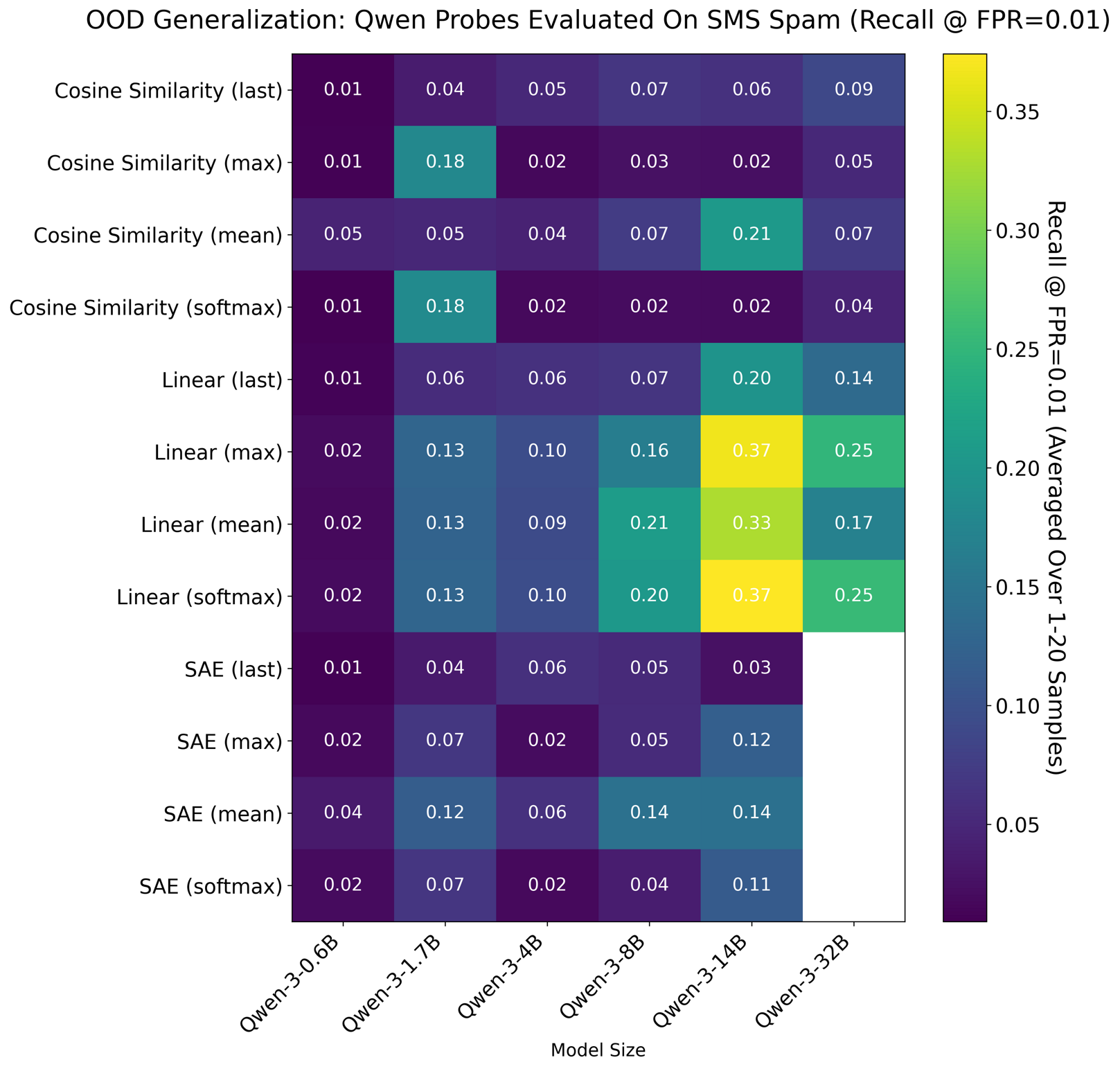}
        \caption{Recall @ 0.01 FPR Heatmap}
    \end{subfigure}
    \caption{Heatmaps showing generalization performance across model sizes and number of positive examples.}
\end{figure}

\end{document}